\documentclass[twoside,11pt]{article}

\usepackage{jmlr2e}

\usepackage{makeidx}  
\usepackage{amsmath}
\usepackage{amsfonts}
\usepackage{amscd}
\usepackage[tableposition=top]{caption}
\usepackage{ifthen}
\usepackage[utf8]{inputenc}
\usepackage{graphicx}
\usepackage{graphicx}
\usepackage{caption}
\usepackage{subcaption}

\usepackage{color}

\usepackage{makeidx}
\usepackage{verbatim}
\usepackage{placeins}

\usepackage{amssymb}
\usepackage{bbm}

\usepackage{multirow} 
\usepackage{makecell}
\usepackage{hhline} 
\usepackage{array} 

\usepackage{siunitx}

\usepackage{flexisym}

\usepackage{algorithm}
\usepackage[noend]{algpseudocode}

\DeclareMathOperator*{\argmax}{arg\,max}

\makeatletter
\def\BState{\State\hskip-\ALG@thistlm}
\makeatother

\definecolor{amethyst}{rgb}{0.6, 0.4, 0.8}
\definecolor{ao}{rgb}{0.0, 0.0, 1.0}
\definecolor{red}{rgb}{1, 0, 0}
\definecolor{green}{rgb}{0.2, 1, 0.2}
\definecolor{blue}{rgb}{0, 0, 1}
\usepackage[usenames,dvipsnames,table]{xcolor}

\def\condindep{\perp\!\!\!\perp}
\def\notcondindep{\not\!\perp\!\!\!\perp}

\jmlrheading{15}{2014}{0-0}{02/15}{0/00}{Lewis P. G. Evans and Niall M. Adams and Christoforos Anagnostopoulos}

\ShortHeadings{Estimating Optimal Active Learning via Model Retraining Improvement}{Evans and Adams and Anagnostopoulos}
\firstpageno{1}

\begin{document}

\title{Estimating Optimal Active Learning via Model Retraining Improvement}

\author{\name Lewis P. G. Evans \email lewis.evans10@imperial.ac.uk \\
       \addr Department of Mathematics\\
       Imperial College London\\
	London, SW7 2AZ, United Kingdom
       \AND
       \name Niall M. Adams \email n.adams@imperial.ac.uk \\
       \addr Department of Mathematics\\
       Imperial College London\\
	London, SW7 2AZ, United Kingdom\\
       \addr
	Heilbronn Institute for Mathematical Research\\
	University of Bristol\\
	PO Box 2495, Bristol, BS8 9AG, United Kingdom
	\AND
       \name Christoforos Anagnostopoulos \email canagnos@imperial.ac.uk \\
       \addr Department of Mathematics\\
       Imperial College London\\
	London, SW7 2AZ, United Kingdom}

\editor{Yoav Freund} 

\maketitle

\begin{abstract}

A central question for active learning (AL) is: ``what is the optimal selection?''
Defining optimality by classifier loss produces a new characterisation of optimal AL behaviour, by treating expected loss reduction as a statistical target for estimation.

This target forms the basis of \emph{model retraining improvement} (MRI), a novel approach providing a statistical estimation framework for AL.
This framework is constructed to address the central question of AL optimality, and to motivate the design of estimation algorithms.

MRI allows the exploration of optimal AL behaviour, and the examination of AL heuristics, showing precisely how they make sub-optimal selections.
The abstract formulation of MRI is used to provide a new guarantee for AL, that an unbiased MRI estimator should outperform random selection.

This MRI framework reveals intricate estimation issues that in turn motivate the construction of new statistical AL algorithms.
One new algorithm in particular performs strongly in a large-scale experimental study, compared to standard AL methods.
This competitive performance suggests that practical efforts to minimise estimation bias may be important for AL applications.

\end{abstract}

\begin{keywords}
  active learning, model retraining improvement, estimation framework, expected loss reduction, classification
\end{keywords}

\section{Introduction}
\label{section:Introduction}

Classification is a central task in statistical inference and machine learning.
In certain cases unlabelled data is plentiful, and a subset can be queried for labelling.
Active learning (AL) seeks to intelligently select this subset of unlabelled examples, to improve a base classifier.
Examples include medical image diagnosis and document categorisation \citep{Dasgupta2008,Hoi2006}.
Many AL methods are heuristic, alongside a few theoretical approaches reviewed by \citet{Settles2009,Olsson2009}.
AL method performance is often assessed by large-scale experimental studies such as \citet{Guyon2011,Kumar2010,Evans2013}.

A prototypical AL scenario consists of a classification problem and a classifier trained on a small labelled dataset.
The classifier may be improved by retraining with further examples, systematically selected from a large unlabelled pool.
This formulation of AL raises the central question for AL, ``what is the optimal selection?''

Performance in classification is judged by loss functions such as those described in \citet{Hand1997}, suggesting that optimality in AL selection should be characterised in terms of classifier loss.
This suggests that the optimal selection should be defined as the example that maximises the expected loss reduction.
This statistical quantity forms the basis of model retraining improvement (MRI), a novel statistical framework for AL.
Compared to heuristic methods, a statistical approach provides strong advantages, both theoretical and practical, described below.


This MRI estimation framework addresses the central question by formally defining optimal AL behaviour. 
Creating a mathematical abstraction of  optimal AL behaviour allows reasoning about heuristics, e.g. showing precisely how they make sub-optimal choices in particular contexts. 
Within this framework, an ideal unbiased MRI estimator is shown to have the property of outperforming random selection, which is a new guarantee for AL.

Crucially, MRI motivates the development of novel algorithms that perform strongly compared to standard AL methods.
MRI estimation requires a series of steps, which are subject to different types of estimation problem.
Algorithms are constructed to approximate MRI, taking different estimation approaches.

A large-scale experimental study evaluates the performance of the two new MRI estimation algorithms, alongside standard AL methods. 
The study explores many sources of variation: classifiers, AL algorithms, with real and abstract classification problems (both binary and multi-class). 
The results show that the MRI-motivated algorithms perform competitively in comparison to standard AL methods.

This work is structured as follows: first the background of classification and AL are described in Section \ref{section:Background}.
Section \ref{section:Example Quality} defines MRI, illustrated by an abstract classification problem in Section \ref{subsection:Theoretical Example to Illustrate Example Quality}.
MRI estimation algorithms are described in Section \ref{section:Algorithms to Estimate Example Quality} and evaluated in a large-scale experimental study of Section \ref{section:Experiments and Results}, followed by concluding remarks.

\section{Background}
\label{section:Background}

The background contexts of classification and AL are described, followed by a brief review of relevant literature, with particular focus on methods that are used later in the paper.

\subsection{Classification}
\label{subsection:Classification}

The categorical response variable $Y$ is modelled as a function of the covariates ${\bf X}$.
For the response $Y$ there are $k$ classes with class labels $\{c_1, c_2, ..., c_k\}$. 
Each classification example is denoted $ ( {\bf x}, y) $, where ${\bf x}$ is a $d$-dimensional covariate vector and $y$ is a class label.
The class prior is denoted $\boldsymbol\pi$.

The Bayes classifier is an idealisation based on the true distributions of the classes, thereby producing optimal probability estimates, and class allocations given a loss function.
Given a covariate vector ${\bf x}$, the Bayes classifier outputs the class probability vector of $Y|{\bf x}$ denoted ${\bf p} = (p_j)_1^k$.
A \emph{probabilistic} classifier estimates the class probability vector as ${\bf \hat{p}} = (\hat{p}_j)_1^k$, and allocates ${\bf x}$ to class $\hat{y}$ using decision theoretic arguments, often using a threshold. 
This allocation function is denoted $h$: $\hat{y} = h({\bf \hat{p}})$.
For example, to minimise misclassification error, the most probable class is allocated: $\hat{y} = h ({\bf \hat{p}}) = \argmax_{j} (\hat{p}_j)$.
The objective of classification is to learn an allocation rule with good generalisation properties.

A somewhat non-standard notation is required to support this work, which stresses the dependence of the classifier on the training data.
A dataset is a set of examples, denoted $D = \{ {\bf x}_i,y_i \}_{i=1}^n$, where $i$ indexes the example.
This indexing notation will be useful later. 
A dataset $D$ may be subdivided into training data $D_T$ and test data $D_E$.
This dataset division may be represented by index sets, for example, $T \cup E = \{1, ..., n\}$, showing the data division into training and test subsets.

First consider a parametric classifier, for example linear discriminant analysis or logistic regression \citep[Chapter~4]{Bishop2007}.
A parametric classifier has estimated parameters $\boldsymbol{\hat{\theta}}$, which can be regarded as a fixed length vector (fixed given $d$ and $k$).
These parameters are estimated by model fitting, using the training data: $\boldsymbol{\hat{\theta}} = \theta (D_T)$, where $\theta()$ is the model fitting function.
This notation is intended to emphasize the dependence of the estimated parameters $\boldsymbol{\hat{\theta}}$ on the training data $D_T$.

Second, this notation is slightly abused to extend to non-parametric classifiers. 
The complexity of non-parametric classifiers may increase with sample size, hence they cannot be represented by a fixed length object.
In this case $\boldsymbol{\hat{\theta}}$ becomes a variable-length object containing the classifier's internal data (for example the nodes of a decision tree, or the stored examples of $K$-nearest-neighbours).

While the contents and meaning of $\boldsymbol{\hat{\theta}}$ would be very different, the classifier's functional roles are identical: model training produces $\boldsymbol{\hat{\theta}}$, which is used to predict class probabilities.
This probability prediction is denoted ${\bf \hat{p}} = \phi(\boldsymbol{\hat{\theta}}, \mathbf{x})$.
These predictions are in turn used to assess classifier performance.

To consider classifier performance, first assume a fixed training dataset $D_T$.
Classifier performance is assessed by a loss function, for example error rate, which quantifies the disagreement between the classifier's predictions and the truth. 
The empirical loss for a single example is defined via a loss function $g(y, {\bf \hat{p}})$.
Many loss functions focus on the allocated class, for example error rate, $g_e(y, {\bf \hat{p}}) = \mathbbm{1} (y \ne h({\bf \hat{p}}))$.
Other loss functions focus on the predicted probability, for example log loss, $g_o({\bf \hat{p}}) = \sum_{j=1}^k ( p_j \textrm{ log } \hat{p}_j )$.

The estimated probabilities ${\bf \hat{p}}$ are highly dependent on the estimated classifier $\boldsymbol{\hat{\theta}}$.
To emphasize that dependence, the empirical loss for a single example is denoted $M(\boldsymbol{\hat{\theta}}, {\bf x}, y) = g(y, {\bf \hat{p}})$.
For example, error rate empirical loss is denoted $M_e(\boldsymbol{\hat{\theta}}, {\bf x}, y)$.

In classification, generalisation performance is a critical quantity. 
For this reason, empirical loss is generalised to expected loss, denoted $L(\boldsymbol{\hat{\theta}})$:
\begin{equation*} 
L (\boldsymbol{\hat{\theta}}) = E_{ {\bf X},Y} [M (\boldsymbol{\hat{\theta}}, {\bf x}, y)] = E_{Y|{\bf X}} E_{\bf X} [M (\boldsymbol{\hat{\theta}}, {\bf x}, y)].
\end{equation*}
This expected loss $L$ is defined as an expectation over all possible test data, given a specific training set.
The expected error rate and log loss are denoted $L_e$ and $L_o$.
Hereafter loss will always refer to the expected loss $L$.
The loss $L(\boldsymbol{\hat{\theta}})$ is dependent on the data $D$ used to train the classifier, emphasized by rewriting $L(\boldsymbol{\hat{\theta}})$ as $L(\theta(D))$ since $\boldsymbol{\hat{\theta}} = \theta(D)$.

The change in the loss as the number of labelled examples increases is of great methodological interest.
This function is known as the \emph{learning curve}, typically defined as the change of expected loss with the number of examples.
Learning curves are illustrated in Figure \ref{figure:Performance Comparison of AL and RS}, and discussed in \citet{Provost2003,Gu2001,Kadie1995}.

\subsection{Active Learning}
\label{subsection:Active Learning}

The context for AL is an abundance of unlabelled examples, with labelled data either expensive or scarce.
Good introductions to AL are provided by \citet{Dasgupta2011}, \citet{Settles2009} and \citet{Olsson2009}.

An algorithm can select a few unlabelled examples to obtain their labels from an oracle (for example a human expert).
This provides more labelled data which can be included in the training data, potentially improving a classifier.
Intuitively some examples may be more informative than others, so systematic example selection should maximise classifier improvement.

In \emph{pool-based} AL, there is an unlabelled pool of data $X_P$ from which examples may be selected for labelling. 
This pool provides a set of examples for label querying, and also gives further information on the distribution of the covariates.
Usually there is also a (relatively small) initial dataset of labelled examples, denoted $D_I$, typically assumed to be iid in AL. 
This work considers the scenario of pool-based AL.

In AL it is common to examine the learning curve, by repeating the AL selection step many times (\emph{iterated AL}).
At each selection step, the loss is recorded, and this generates a set of losses, which define the learning curve for the AL method.
Iterated AL allows the exploration of performance over the learning curve, as the amount of labelled data grows.
This repeated application of AL selection is common in both applications and experimental studies \citep{Guyon2011,Evans2013}.

In contrast to iterated AL, the AL selection step may occur just once (\emph{single-step AL}).
The question of iterated or single-step AL is critical, because iterated AL inevitably produces covariate bias in the labelled data.
The covariate bias from iterated AL creates a selection bias problem, which is intrinsic to AL. 

At each selection step, an AL method may select a single example from the pool (\emph{individual AL}) or several examples at once (\emph{batch AL}).
AL applications are often constrained to use batch AL for pragmatic reasons \citep{Settles2009}.

Turning to AL performance, consider random selection (RS) where examples are chosen randomly (with equal probability) from the pool.
By contrast, AL methods select some examples in preference to others.
Under RS and AL, the classifier receives exactly the same number of labelled examples; thus RS provides a reasonable benchmark for AL \citep{Guyon2011,Evans2013}.
The comparison of methods to benchmarks is available in experiments but not in real AL applications \citep{Provost2010}.

\begin{figure}
\centering
\includegraphics[scale=0.8]{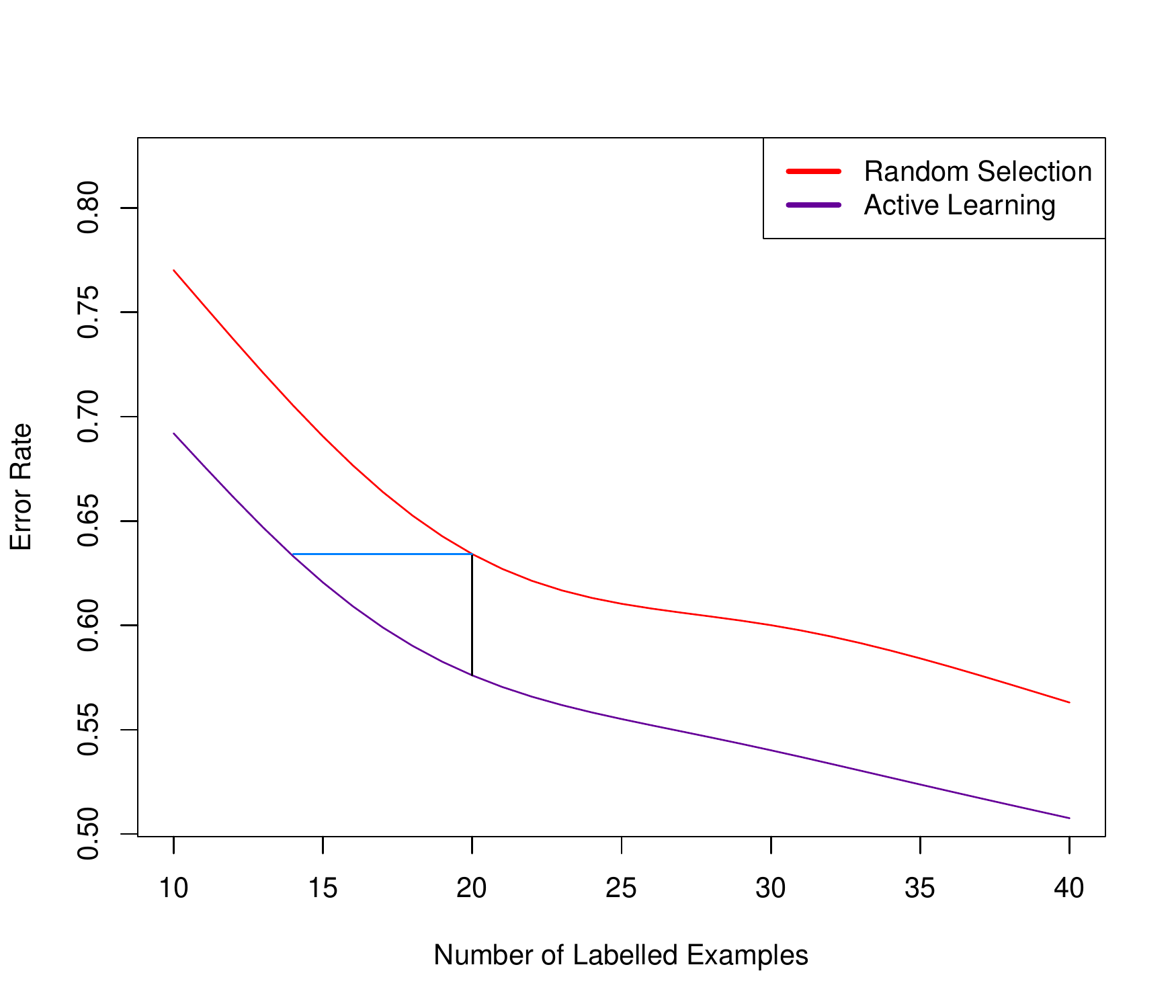}
\caption{Performance comparison of active learning and random selection, showing that a classifier often improves faster under AL than under RS.
In both cases the loss decreases as the number of labelled examples increases; however, AL improves faster than RS. 
These curves are smoothed averages from multiple experiments. 
The black vertical line illustrates the \emph{fixed-label comparison}, whereas the blue horizontal line shows the \emph{fixed-loss comparison } (see Section \ref{subsection:Active Learning}). 
The classification problem is ``Abalone'' from UCI, a three-class problem, using classifier $5$-nn, and Shannon entropy as the AL method.
}
\label{figure:Performance Comparison of AL and RS}
\end{figure}

Classifier performance should improve, at least on average, even under the benchmark RS, since the classifier receives more training data (an issue explored below).
AL performance assessment should consider how much AL outperforms RS. 
Hence AL performance addresses the relative improvement of AL over RS, and the relative ranks of AL methods, rather than the absolute level of classifier performance.
Figure \ref{figure:Performance Comparison of AL and RS} shows the losses of AL and RS as the number of labelled examples increases.

Figure \ref{figure:Performance Comparison of AL and RS} shows two different senses in which AL outperforms RS: first AL achieves better loss reduction for the same number of labels (\emph{fixed-label comparison}), and second AL needs fewer labels to reach the same classifier performance (\emph{fixed-loss comparison}).
Together the fixed-label comparison and fixed-loss comparison form the two fundamental aspects of AL performance.
The fixed-label comparison first fixes the number of labels, then seeks to minimise loss.
Several established performance metrics focus on the fixed-label comparison: AUA, ALC and WI \citep{Guyon2011,Evans2013}.
The fixed-label comparison is more common in applications where the costs of labelling are significant \citep{Settles2009}.

Under the fixed-loss comparison, the desired level of classifier loss is fixed, the goal being to minimise the number of labels needed to reach that level.
Label complexity is the classic example, where the desired loss level is a fixed ratio of asymptotic classifier performance \citep{Dasgupta2011}.
Label complexity is often used as a performance metric in contexts where certain assumptions permit analytically tractable results, for example \citet{Dasgupta2011}.

\subsection{Overview of Active Learning Methods}
\label{subsection:Literature Review}

A popular AL approach is the uncertainty sampling heuristic, where examples are chosen with the greatest class uncertainty \citep{Thrun1992,Settles2009}.
This approach selects examples of the greatest classifier uncertainty in terms of class membership probability.
The idea is that these uncertain examples will be the most useful for tuning the classifier's decision boundary.
Example methods include Shannon entropy (SE), least confidence and maximum uncertainty.
For a single unlabelled example {\bf x}, least confidence is defined as
$ U_L ({\bf x}, \theta(D)) = 1 - \hat{p}(\hat{y} | {\bf x})$,
where $\hat{p}(\hat{y} | {\bf x})$ is the classifier's estimated probability of the allocated class $\hat{y}$.
Shannon entropy is defined as
$ U_E ({\bf x}, \theta(D)) = \sum_{j=1}^k \hat{p}_j \, \textrm{log} (\hat{p}_j)$.
The uncertainty sampling approach is popular and efficient, but lacks theoretical justification.

Version space search is a theoretical approach to AL, where the version space is the set of hypotheses (classifiers) that are consistent with the data \citep{Mitchell1997,Dasgupta2011}.
Learning is then interpreted as a search through version space for the optimal hypothesis.
The central idea is that AL can search this version space more efficiently than RS.


Query by committee (QBC) is a heuristic approximation to version space search \citep{Seung1992}.
Here a committee of classifiers is trained on the labelled data, which then selects the unlabelled examples where the committee's predictions disagree the most.
This prediction disagreement may focus on predicted classes (for example vote entropy) or predicted class probabilities (for example average Kullback-Leibler divergence); see \citet{Olsson2009}.
These widely used versions of QBC are denoted QbcV and QbcA.
A critical choice for QBC is the classifier committee, which lacks theoretical guidance.
In this sense version space search leaves the optimal AL selection unspecified.

Another approach to AL is exploitation of cluster structure in the pool.
Elucidating the cluster structure of the pool could provide valuable insights for example selection.
\citet{Dasgupta2011} gives a motivating example: if the pool clusters neatly into $b$ class-pure clusters where $b=k$, then $b$ labels could suffice to build an optimal classifier.
This very optimistic example does illustrate the potential gain.

A third theoretical approach, notionally close to our contribution, is error reduction, introduced in \citet{Roy2001}.
This approach minimises the loss of the retrained classifier, which is the loss of the classifier which has been retrained on the selected example.
Roy and McCallum consider two loss functions, error rate and log loss, to construct two quantities, which are referred to here as expected future error (EFE) and expected future log loss (EFLL).
Those authors focus on methods to estimate EFE and EFLL, before examining the experimental performance of their estimators.


Given a classifier fitting function $\theta$, labelled data $D$ and a single unlabelled example ${\bf x}$, EFE is defined as 
\begin{equation*} 
EFE({\bf x}, \theta, D) = - E_{Y | {\bf x}} [ L_e(\theta(D \cup ({\bf x},Y)) ] = - \sum_{j=1}^k \{ p_j \, L_e(\theta(D \cup ({\bf x},c_j)) \},
\end{equation*}
where $L_e$ is error rate (see Section \ref{subsection:Classification}).
EFLL is defined similarly to EFE, with log loss $L_o$ replacing error rate $L_e$.
Both of these quantities average over the unobserved label $Y|{\bf x}$.

Roy and McCallum define an algorithm to calculate EFE, denoted EfeLc, which approximates the loss using the unlabelled pool for efficiency.
Specifically it approximates error rate $L_e$ by the total least confidence over the entire pool:
\begin{equation*} 
L_e(\theta(D)) \approx \sum_{ {\bf x}_i \in X_P} U_L({\bf x}_i, \theta(D)),
\end{equation*}
where $X_P$ are the unlabelled examples in the pool.
The uncertainty function $U_L$ is intended to capture the class uncertainty of an unlabelled example.

Roy and McCallum propose the following approximation for the value of EFE by calculating
\begin{equation} \label{eq:efelc}
\begin{split}
f_1({\bf x}, \theta, D) = - \sum_{j=1}^k \left\{ \hat{p}_j \sum_{ {\bf x}_i \in X_P} U_L({\bf x}_i, \theta(D \cup ({\bf x}_i,c_j))) \right\} 
= - \sum_{j=1}^k \left\{ \hat{p}_j \sum_{ {\bf x}_i \in X_P} \left( 1 - \hat{p}(\hat{y_i} | {\bf x}_i) \right) \right\}.
\end{split}
\end{equation}
Here $\hat{p}_j$ is the current classifier's estimate of the class probability for class $j$, while $\hat{y}_i$ is the predicted label for $\mathbf{x}_i$ after a training update with the example $(\mathbf{x}, c_j)$.
Note that EfeLc uses the the classifier's posterior estimates \emph{after} an update (to estimate the loss), whereas the uncertainty sampling approaches use the \emph{current} classifier's posterior estimates (to assess uncertainty).

This approximation of $L_e$ by the total least confidence over the pool is potentially problematic.
It is easy to construct cases (for example an extreme outlier) where a labelled example would reduce a classifier's uncertainty, but also increase the overall error; such examples call into question the approximation of error by uncertainty.
In the absence of further assumptions or motivation, it is hard to anticipate the statistical properties of $f_1$ in Equation \ref{eq:efelc} as an estimator.
Further, EfeLc uses the same data to train the classifier and to estimate the class probabilities, thereby risking bias in the estimator (an issue explored further in Section \ref{section:Algorithms to Estimate Example Quality}).

The error reduction approach is similar in spirit to MRI, since the optimal example selection is first considered, and then specified in terms of classifier loss.
In that sense, the quantity EFE is a valuable precursor to model retraining improvement, which is defined later in Equation \ref{eq:eqc}.
However EFE omits the loss of the current classifier, which proves important when examining improvement (see Section \ref{subsection:Theoretical Example to Illustrate Example Quality}).
Further, EFE is only defined for individual AL, while MRI defines targets for both batch and individual AL.

The estimation of a statistical quantity, consisting of multiple components, raises several statistical choices, in terms of component estimators and how to use the data.
These choices are described and explored in Section \ref{section:Algorithms to Estimate Example Quality}, whereas Roy and McCallum omit these choices, providing just a single algorithmic approach.
In that sense, Roy and McCallum do not use EFE to construct an estimation framework for algorithms.
Nor do Roy and McCallum use EFE to examine optimal AL behaviour, or compare it to the behaviour of known AL methods; Section \ref{subsection:Theoretical Example to Illustrate Example Quality} provides such an examination and comparison using MRI.
Finally, the EFE algorithms do not show strong performance in the experimental results of Section \ref{section:Experiments and Results}.


The current literature does not provide a statistical estimation framework for AL; MRI addresses this directly in Section \ref{section:Example Quality}.

\section{Model Retraining Improvement}
\label{section:Example Quality}

Here the statistical target, model retraining improvement, is defined and motivated as an estimation target, both theoretically and for applications.
This further lays the groundwork for MRI as a statistical estimation framework for AL.
This Section defines the statistical target as an expectation, while Section \ref{section:Algorithms to Estimate Example Quality} describes estimation problems, and algorithms for applications.


\subsection{The Definition of Model Retraining Improvement}
\label{subsection:The Definition of Example Quality}

\begin{table}[h!b!p!] 
\caption{
Notation.
}
\label{table:Notation 1}
\centering
\def\arraystretch{1.2}%
\begin{tabular}{|c|c|}
\hline
\multicolumn{2}{c}{Notation}\\
\hline
\hline
Symbol & Description \\
\hline
\hline
\makecell{ $(\mathbf{X}, Y)$ } & Underlying distribution of the classification problem \\
\hline
\makecell{$\mathbf{p}$} & Bayes class probability vector, for covariate $\mathbf{x}$: $\mathbf{p} = p(Y|\mathbf{x}) = \{ p(c_j|\mathbf{x}) \}_{j=1}^k$\\
\hline
\makecell{$\theta$} & Classifier training function \\
\hline
\makecell{$\boldsymbol{\hat{\theta}}$} & Classifier estimated parameters, where $\boldsymbol{\hat{\theta}} = \theta(D_T)$\\
\hline
\makecell{$\phi$} & Classifier prediction function; class probability vector $\mathbf{\hat{p}} = \phi(\boldsymbol{\hat{\theta}}, \mathbf{x})$ \\
\hline
\makecell{$D_S$} & The labelled data: $D_S = (X_S, Y_S) = \{ \mathbf{x}_i, y_i \}^{i \in S}$\\
\hline
\makecell{$X_P$} & The unlabelled pool \\
\hline
\makecell{$Q^c$} & Statistical target, optimal for individual AL \\
\hline
\makecell{$B^c$} & Statistical target, optimal for batch AL \\
\hline
\makecell{$L$} & Classifier loss \\
\hline
\makecell{${L '}_j$} & Classifier future loss, after retraining on $(\mathbf{x}, c_j)$: ${L '}_j = L(\theta(D_S \cup (\mathbf{x}, c_j))$\\
\hline
\makecell{$\mathbf{L '}$} & Classifier future loss vector, for covariate $\mathbf{x}$: $\mathbf{L '} = \{ L(\theta(D_S \cup (\mathbf{x}, c_j)) \}_{j=1}^k$\\
\hline
\hline
\end{tabular}
\end{table}

The notation is summarised in Table \ref{table:Notation 1}.
To define the statistical target, expectations are formed with respect to the underlying distribution $({\bf X}, Y)$.
Assume a fixed dataset $D_S$ sampled i.i.d. from the joint distribution $({\bf X}, Y)$.
The dependence of the classifier $\boldsymbol{\hat{\theta}}$ on the data $D_S$ is critical, with the notation $\boldsymbol{\hat{\theta}} = \theta(D_S)$ intended to emphasize this dependence.

First assume a base classifier already trained on a dataset $D_S$.
Consider how much a single labelled example improves performance.
The single labelled example $({\bf x},y)$ will be chosen from a \emph{labelled} dataset $D_W$. 
The loss from retraining on that single labelled example is examined in order to later define the loss for the expected label of an unlabelled example.

Examine the selection of a single labelled example $({\bf x},y)$ from $D_W$, given the labelled data $D_S$, the classifier training function $\theta$ and a loss function $L$.	
The reduction of the loss for retraining on that example is defined as actual-MRI, denoted $Q^a$:

\begin{equation*} 
Q^a({\bf x}, y, \theta, D_S) = L(\theta(D_S)) - L(\theta(D_S \cup ({\bf x},y)). 
\end{equation*}

$Q^a$ is the actual classifier improvement from retraining on the labelled example $({\bf x},y)$.
The goal here is to maximise the reduction of loss.
The greatest loss reduction is achieved by selecting the example $({\bf x_*},y_*)$ from $D_W$ that maximises $Q^a$, given by
\begin{equation*} 
{(\bf x_*}, y_*) = \argmax_{ ({\bf x}, y) \in D_W} Q^a({\bf x}, y, \theta, D_S).
\end{equation*} 


Turning to AL, the single example ${\bf x}$ is unlabelled, and will be chosen from the unlabelled pool $X_P$.
Here the unknown label of ${\bf x}$ is a random variable, $Y|{\bf x}$, and taking its expectation allows the expected loss to defined, this being the classifier loss after retraining with the unlabelled example and its unknown label.
Thus the expected loss is defined using the expectation over the label $Y|{\bf x}$ to form conditional-MRI, denoted $Q^c$:

\begin{equation} \label{eq:eqc}
\begin{split}
Q^c({\bf x}, \theta, D_S) = E_{Y | {\bf x}} [Q^a({\bf x}, Y, \theta, D_S)] = L(\theta(D_S)) - E_{Y | {\bf x}} [ L(\theta(D_S \cup ({\bf x},Y)) ] \\
= L(\theta(D_S)) - \sum_{j=1}^k \{ p_j \, L(\theta(D_S \cup ({\bf x},c_j)) \}
= \underbrace{ L(\theta(D_S)) }_{\text{Term } T_c} - \underbrace{ \sum_{j=1}^k p_j \, {L '}_j }_{\text{Term } T_e} = L(\theta(D_S)) - \mathbf{p} \cdot \mathbf{L '},
\end{split}
\end{equation}

where $\mathbf{p}$ denotes the Bayes class probability vector $p(Y|{\bf x})$.
${L '}_j$ denotes a single future loss, from retraining on $D_S$ together with one example ${\bf x}$ given class $c_j$.
$\mathbf{L '}$ denotes the future loss vector, i.e. the vector of losses from retraining on $D_S$ together with one example, that example being ${\bf x}$ combined with each possible label $c_j$:
 $\mathbf{L '} = \{ {L '}_j \}_{j=1}^k = \{ L(\theta(D_S \cup ({\bf x}, c_j)))\}_{j=1}^k$.

Term $T_c$ is the loss of the current classifier, given the training data $D_S$.
Term $T_e$ is the expected future loss of the classifier, after retraining on the enhanced dataset $(D_S \cup (\mathbf{x}, Y|\mathbf{x}))$.
$Q^c$ is defined as the difference between Terms $T_c$ and $T_e$, i.e. the difference between the current loss and the expected future loss.
Thus $Q^c$ defines the expected loss reduction, from retraining on the example ${\bf x}$ with its unknown label. 
In this sense $Q^c$ is an improvement function, since it defines exactly how much this example will improve the classifier.

The unlabelled example ${\bf x_*}$ from the pool $X_P$ that maximises $Q^c$ is the optimal example selection:
\begin{equation} \label{eq:maximise_eqc}
{\bf x_*} = \argmax_{ {\bf x} \in X_P} Q^c({\bf x}, \theta, D_S).
\end{equation}

Novel algorithms are constructed to estimate the target $Q^c$, given in Section \ref{section:Algorithms to Estimate Example Quality}.

For an abstract classification problem, the target $Q^c$ can be evaluated exactly, to reveal the best and worst possible loss reduction, by maximising and minimising $Q^c$.
Figure \ref{figure:Maximum and minimum AL performance with simulated data, exact EQ} shows that the best and worst AL performance curves are indeed obtained by maximising and minimising $Q^c$.

The statistical quantity $Q^c$ defines optimal AL behaviour for any dataset $D_S$, whether iid or not, including the case of iterated AL, which generates a covariate bias in $D_S$ (see Section \ref{subsection:Active Learning}).
Given $Q^c$ for the selection of a single example, i.e. for individual AL, the optimal behaviour is now extended to batch AL, the selection of multiple examples, via the target $B^c$, given below.

\begin{figure}
\centering
\includegraphics[scale=0.8]{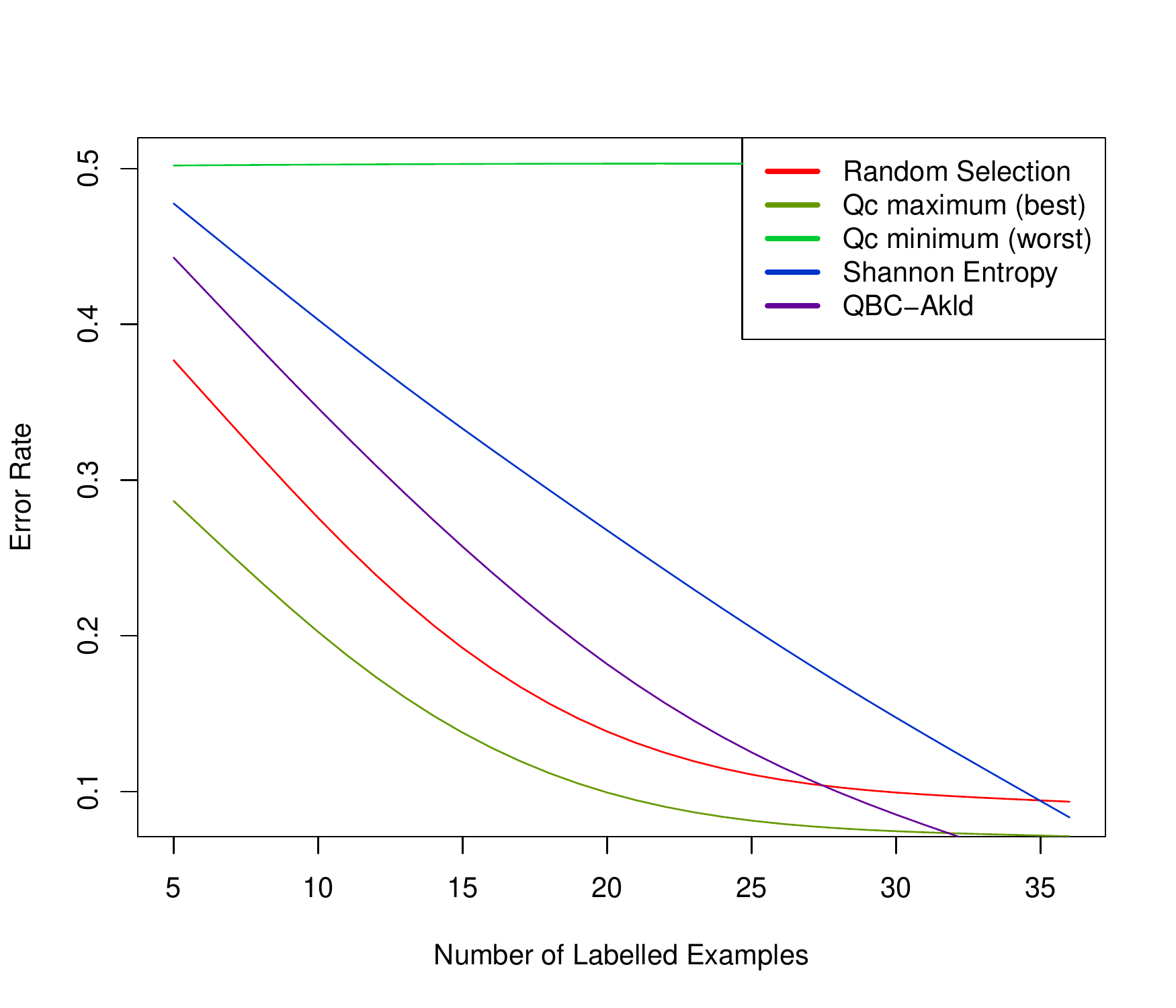}
\caption{The best and worst AL performance curves are obtained by maximising and minimising the target $Q^c$, which demonstrate the extremes of AL performance.
With simulated data, $Q^c$ can be calculated exactly; here the classification problem is the Four-Gaussian problem (illustrated in Figure \ref{fig:Ripley Four-Gaussian Problem}).
These curves are smoothed from multiple experiments, using the classifier $5$-nn. 
}
\label{figure:Maximum and minimum AL performance with simulated data, exact EQ}
\end{figure}

\subsubsection{Model Retraining Improvement for Batch Active Learning}
\label{subsubsection:Example Quality for Batch Active Learning}

In batch AL, multiple examples are selected from the pool in one selection step.
Each chosen batch consists of $r$ examples.
Here MRI provides the statistical target $B^c$, the batch improvement function, defined as the expected classifier improvement over an unknown set of labels.

First examine a \emph{fully labelled} dataset $({\bf x}_R, {\bf y}_R)$, where $R$ denotes the index set $\{ {1, ..., r} \}$.
For that fully labelled dataset, the actual loss reduction is denoted $B^a$: 
\begin{equation*} 
B^a({\bf x}_R, {\bf y}_R, \theta, D_S) = L(\theta(D_S)) - L(\theta(D_S \cup ({\bf x}_R,{\bf y}_R)).
\end{equation*}

Second consider the AL context, with a set of \emph{unlabelled} examples ${\bf x_R}$, which is a single batch of examples selected from the pool.
The expected loss reduction for this set of examples is denoted $B^c$: 
\begin{equation*} 
\begin{split}
B^c({\bf x}_R, \theta, D_S) = E_{{\bf Y}_R | {\bf x}_R} [B^a({\bf x}_R, {\bf Y}_R, \theta, D_S)] = L(\theta(D_S)) - E_{{\bf Y}_R | {\bf x}_R} [ L(\theta(D_S \cup ({\bf x}_R,{\bf Y}_R)) ] \\
= L(\theta(D_S)) - \sum_{j_1=1}^k \sum_{j_2=1}^k ... \sum_{j_r=1}^k \{ p_{j_1} p_{j_2} ... p_{j_r} \times L(\theta(D_S \cup ({\bf x_1},c_{j_1}) \cup ({\bf x_2},c_{j_2}) ... \cup ({\bf x_r},c_{j_r})) \}.
\end{split}
\end{equation*}
This expected loss reduction $B^c$ is an expectation taken over the unknown set of labels $({\bf Y}_R | {\bf x}_R)$.
$B^c$ is the statistical target for batch AL, and the direct analog of $Q^c$ defined in Equation \ref{eq:eqc}.

Estimating $B^c$ incurs two major computational costs, in comparison to $Q^c$ estimation.
First there is the huge increase in the number of selection candidates.
For individual AL, each selection candidate is a single example, and there are only $n_p$ candidates to consider (where $n_p$ is the pool size).
Under batch AL, each selection candidate is a set of examples, each set having size $r$; the number of candidates jumps to $n_p \choose r$.
Thus batch AL generates a drastic increase in the number of selection candidates, from $n_p$ to $n_p \choose r$, which presents a major computational cost.

The second cost of $B^c$ estimation lies in the number of calculations per selection candidate.
In individual AL, each candidate requires one classifier retraining and one loss evaluation per class, for all $k$ classes.
However in batch AL, each candidate requires multiple classifier retraining and loss evaluations, each candidate now requiring $k^r$ calculations.
Hence the number of calculations increases greatly, from $k$ to $k^r$, which is a severe computational cost.

%

These major computational costs make direct estimation of the target $B^c$ extremely challenging.
Thus for batch AL the more practical option is to recommend algorithms that estimate $Q^c$, such as the algorithms given in Section \ref{section:Algorithms to Estimate Example Quality}.

$Q^c$ and $B^c$ together define the optimal AL behaviour for individual AL and batch AL.
These targets provide optimal AL behaviour for both single-step and iterated AL. 
The rest of this work focusses on the target $Q^c$ as the foundation of MRI's estimation framework for AL.

\subsection{Abstract Example}
\label{subsection:Theoretical Example to Illustrate Example Quality}

An example using an abstract classification problem is presented, to illustrate MRI in detail.
The stochastic character of this problem is fully specified, allowing exact calculations of the loss $L$, and the statistical target $Q^c$ as functions of the univariate covariate $x$.
To reason about $Q^c$ as a function of $x$, an infinite pool is assumed, allowing any $x \in \mathbb{R}$ to be selected.
These targets are then explored as functions of $x$, and the optimal AL selection $x_*$ is examined (see Equation \ref{eq:maximise_eqc}).

The full stochastic description allows examination of the AL method's selection, denoted $x_r$, and comparison to the optimal selection $x_*$.
This comparison is made below for the popular AL heuristic Shannon entropy, and for random selection.



Imagine a binary univariate problem, defined by a balanced mixture of two Gaussians: $ \{ {\boldsymbol\pi} = (\frac{1}{2}, \frac{1}{2}), (X|Y=c_1) \sim \textrm{N}(-1,1), (X|Y=c_2) \sim \textrm{N}(1,1) \}$.
The true means are denoted ${\mu}_1 = -1, {\mu}_2 = 1$.
The loss function is error rate $L_e$ (defined in Section \ref{subsection:Classification}), while the true decision boundary to minimise error rate is denoted $t = \frac{1}{2} ({\mu}_1 + {\mu}_2)$.

Every dataset $D$ of size $n$ sampled from this problem is assumed to split equally into two class-pure subsets $D_j = \{ y_i = c_j, (x_i, y_i) \in D \}$ each of size $n_j = \frac{n}{2}$; this is sampling while holding the prior fixed.


Consider a classifier that estimates only the class-conditional means, given the true prior $\boldsymbol\pi$ and the true common variance of 1.
The classifier parameter vector is $\boldsymbol{\hat{\theta}} = ( \hat{\mu}_1, \hat{\mu}_2 )$, where $\hat{\mu}_j$ is the sample mean for class $c_j$.
This implies that the classifier's estimated decision boundary to minimise error rate is denoted $\hat{t} = \frac{1}{2} (\hat{\mu}_1 + \hat{\mu}_2)$.


\subsubsection{Calculation and Exploration of $Q^c$}
\label{subsubsection:Calculation of Qc}

Here $Q^c$ is calculated, then explored as a function of $x$. 
The classifier's decision rule $r_1(x)$ minimises the loss $L_e(\boldsymbol{\hat{\theta}})$, and is given in terms of a threshold on the estimated class probabilities by
\begin{displaymath}
   r_1(x) = \left\{
     \begin{array}{lr}
       \hat{y} = c_1 & : \hat{p}_1(x) > \frac{1}{2}, \\
       \hat{y} = c_2 & : \hat{p}_1(x) < \frac{1}{2}, \\
     \end{array}
   \right.
\end{displaymath}
or equivalently, in terms of a decision boundary on $x$, by
\begin{displaymath}
   r_2(x) = \left\{
     \begin{array}{lr}
	\hat{\mu}_1 < \hat{\mu}_2 & : \hat{y} = c_1 \textrm{ if } x < \hat{t}, c_2 \textrm{ otherwise}, \\
	\hat{\mu}_1 > \hat{\mu}_2 & : \hat{y} = c_1 \textrm{ if } x > \hat{t}, c_2 \textrm{ otherwise}. \\
     \end{array}
   \right.
\end{displaymath}


The classifier may get the estimated class means the wrong way around, in the unlikely case that $\hat{\mu}_1 > \hat{\mu}_2$.
As a result the classifier's behaviour is very sensitive to the condition $(\hat{\mu}_1 > \hat{\mu}_2)$, as shown by the second form of the decision rule $r_2(x)$, and by the loss function in Equation \ref{eq:le_equation_1}.

It is straightforward to show that the loss $L_e(\boldsymbol{\hat{\theta}})$ is given by
\begin{equation} \label{eq:le_equation_1}
\begin{split}
L_e(\boldsymbol{\hat{\theta}}) = \frac{1}{2} \{ 1 - F_1(\hat{t}) + F_2(\hat{t}) + \mathbbm{1}(\hat{\mu}_1 > \hat{\mu}_2) [2 F_1(\hat{t}) - 2 F_2(\hat{t})] \},
\end{split}
\end{equation}
where $F_j(x)$ denotes the cdf for class-conditional distribution $(X|Y=c_j)$.

In individual AL an unlabelled point $x$ is chosen for the oracle to label, before retraining the classifier.
Retraining the classifier with a single new example $(x, c_j)$ yields a new parameter estimate denoted $\boldsymbol{\hat{\theta}}^{\prime}_j$, where the mean estimate for class $c_j$ has a new value denoted $\hat{\mu}^{\prime}_j$, with a new estimated boundary denoted $\hat{t}^{\prime}_j$.

Here $\hat{\mu}^{\prime}_j = (1-z)\hat{\mu}_j + z x$ where $z = \frac{2}{n+2}$, $z$ being an updating constant which reflects the impact of the new example on the mean estimate $\hat{\mu}_j$.

To calculate $Q^c$ under error loss $L_e$, observe that the Term $T_e$ from Equation \ref{eq:eqc} is $[ p_1 L_e(\boldsymbol{\hat{\theta}}^{\prime}_1) + p_2 L_e(\boldsymbol{\hat{\theta}}^{\prime}_2) ]$.
Term $T_c$ in Equation \ref{eq:eqc} is directly given by Equation \ref{eq:le_equation_1}.
From Equations \ref{eq:eqc} and \ref{eq:le_equation_1}, $Q^c (x, \theta, D) = L_e(\boldsymbol{\hat{\theta}}) - [ p_1 L_e(\boldsymbol{\hat{\theta}}^{\prime}_1) + p_2 L_e(\boldsymbol{\hat{\theta}}^{\prime}_2) ]$, hence
\begin{equation*} 
\begin{split}
Q^c (x, \theta, D) = \frac{1}{2} \{ 1 - F_1(\hat{t}) + F_2(\hat{t}) + \mathbbm{1}(\hat{\mu}_1 > \hat{\mu}_2) [2 F_1(\hat{t}) - 2 F_2(\hat{t})] \} \\
 - \frac{p_1}{2} \{ 1 - F_1(\hat{t}^{\prime}_1) + F_2(\hat{t}^{\prime}_1) + \mathbbm{1}(\hat{\mu}^{\prime}_1 > \hat{\mu}_2) [2 F_1(\hat{t}^{\prime}_1) - 2 F_2(\hat{t}^{\prime}_1)] \} \\
 - \frac{p_2}{2} \{ 1 - F_1(\hat{t}^{\prime}_2) + F_2(\hat{t}^{\prime}_2) + \mathbbm{1}(\hat{\mu}_1 > \hat{\mu}^{\prime}_2) [2 F_1(\hat{t}^{\prime}_2) - 2 F_2(\hat{t}^{\prime}_2)] \},
\end{split}
\end{equation*}
where $p_j$, $\hat{\mu}^{\prime}_j$, and $\hat{t}^{\prime}_j$ are functions of $x$.

Even for this simple univariate problem, $Q^c(x, \theta, D)$ is a complicated non-linear function of $x$.
Given this complication, $Q^c$ is explored by examining specific cases of the estimated parameter $\boldsymbol{\hat{\theta}}$, shown in Figure \ref{fig:Qc for specific thetaHat cases, Example 1}.
In each specific case of $\boldsymbol{\hat{\theta}}$, $x_*$ yields greatest correction to $\boldsymbol{\hat{\theta}}$ in terms of moving the estimated boundary $\hat{t}$ closer to the true boundary $t$. 
This is intuitively reasonable since error rate is a function of $\hat{t}$ and minimised for $\hat{t} = t$.

In the first two cases (Figures \ref{fig:Qc for specific thetaHat cases, Example 1_case_mu1h=[-0_5]__mu2h=[1_5]} and \ref{fig:Qc for specific thetaHat cases, Example 1_case_mu1h=[-0_9]__mu2h=[1_1]}), the estimated threshold is greater than the true threshold, $\hat{t} > t$.
In these two cases, $x_*$ is negative, hence retraining on $x_*$ will reduce the estimated threshold $\hat{t}$, bringing it closer to the true threshold $t$, thereby improving the classifier.
In the third case (Figure \ref{fig:Qc for specific thetaHat cases, Example 1_case_mu1h=[-1_1]__mu2h=[1_1]}), $\hat{t} = t$ and here the classifier's loss $L_e$ cannot be reduced, shown by $Q^c(x) < 0$ for all $x$. 
The fourth case (Figure \ref{fig:Qc for specific thetaHat cases, Example 1_case_mu1h=[1]__mu2h=[-1]}) is interesting because the signs of the estimated means are reversed compared to the true means, and here the most non-central $x$ offer greatest classifier improvement.
Together these cases show that even for this toy example, the improvement function $Q^c$ is complicated and highly dependent on the estimated parameters.

\begin{figure}
        \centering
        \begin{subfigure}[b]{0.45\textwidth}
                \centering
               \includegraphics[width=\textwidth]{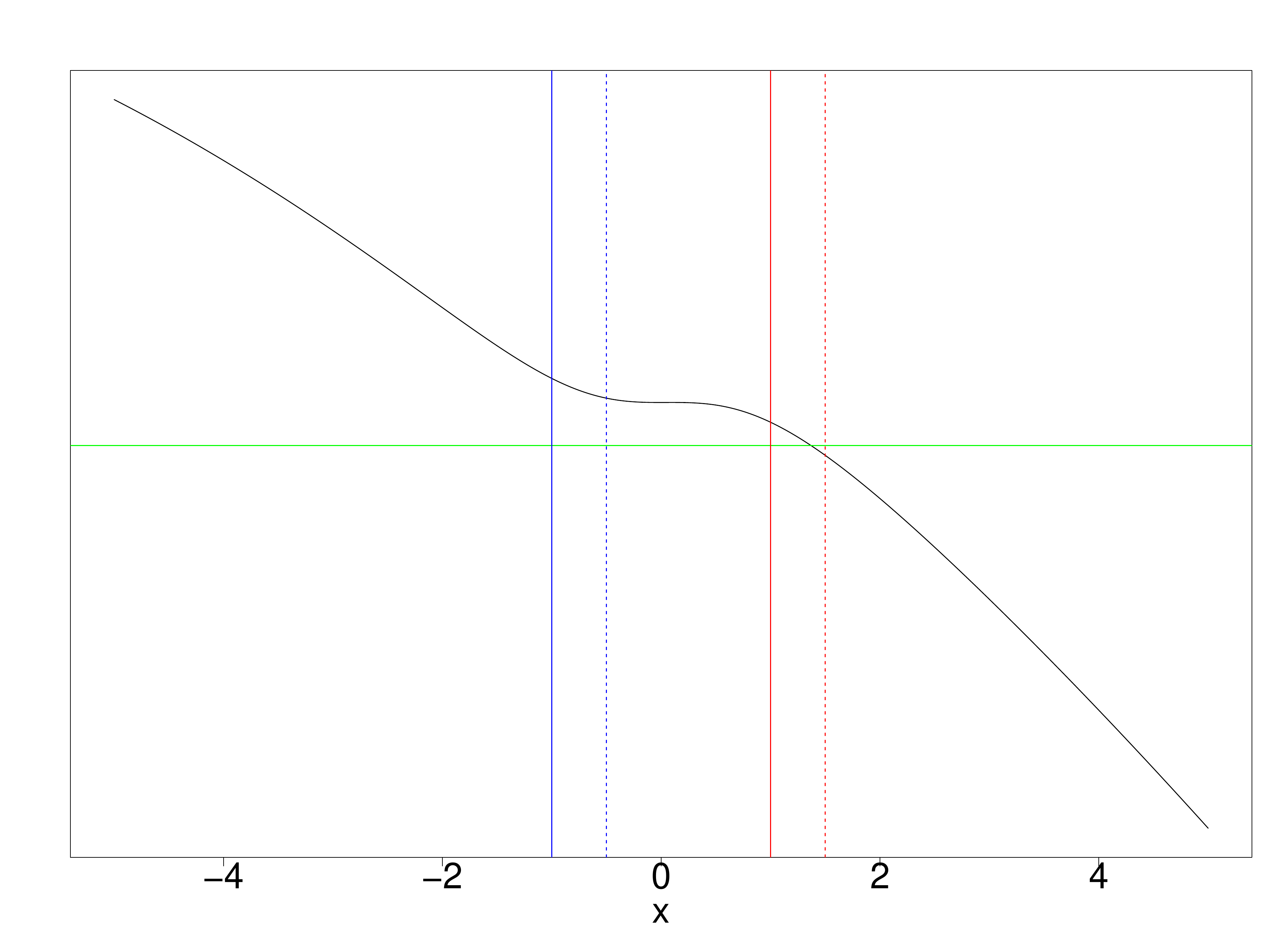}
                \caption{$\boldsymbol{\hat{\theta}} = (\hat{\mu}_1=-0.5,\hat{\mu}_2=1.5)$; \\
$\hat{\mu}_1$, $\hat{\mu}_2$ are right-shifted, $\hat{\mu}_j = {\mu}_j + 0.5$ \\}
                \label{fig:Qc for specific thetaHat cases, Example 1_case_mu1h=[-0_5]__mu2h=[1_5]}
        \end{subfigure}
        \begin{subfigure}[b]{0.45\textwidth}
                \centering
               \includegraphics[width=\textwidth]{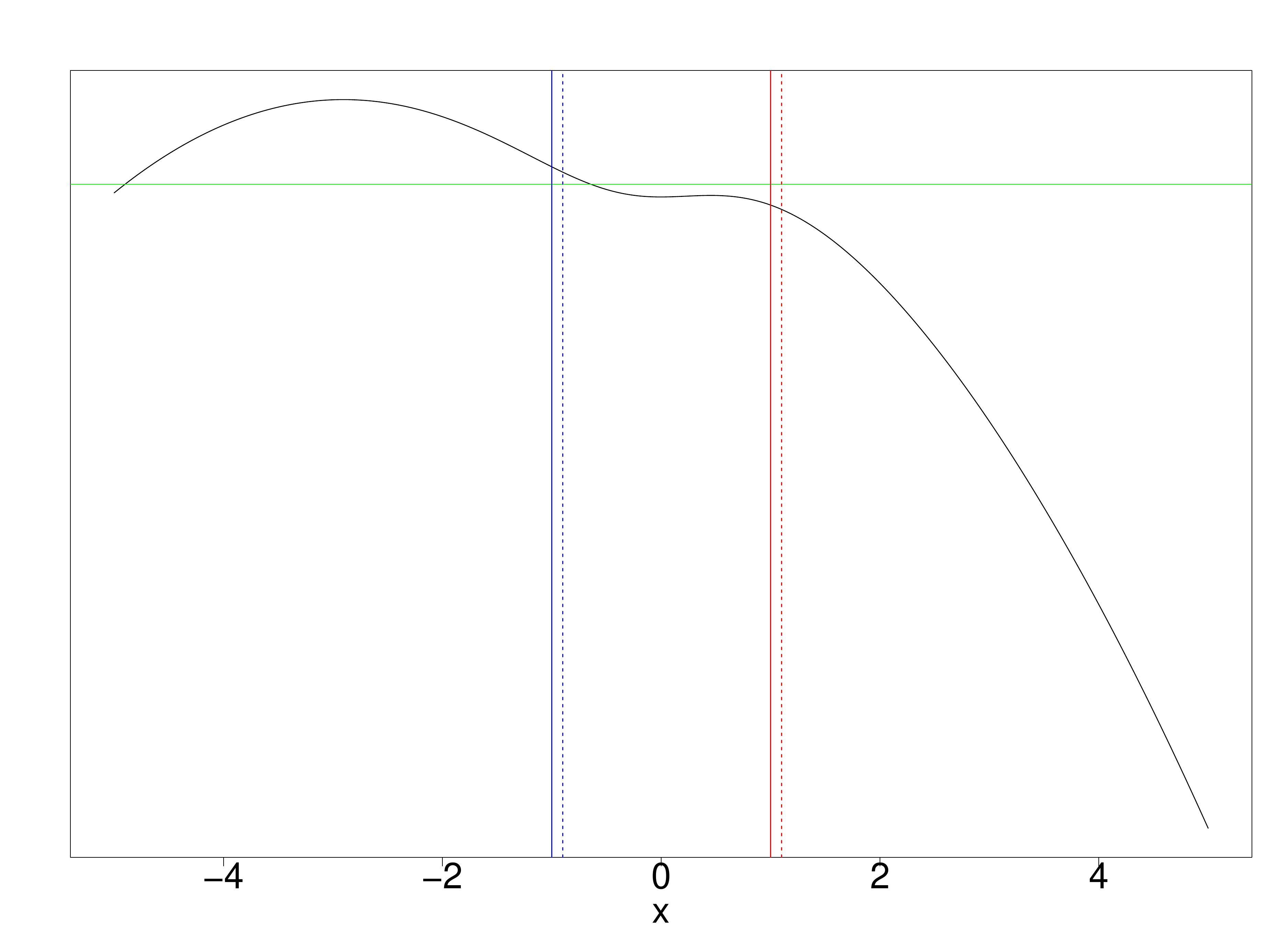}
                \caption{$\boldsymbol{\hat{\theta}} = (\hat{\mu}_1=-0.9,\hat{\mu}_2=1.1)$; \\
$\hat{\mu}_1$, $\hat{\mu}_2$ are right-shifted, $\hat{\mu}_j = {\mu}_j + 0.1$ \\}
                \label{fig:Qc for specific thetaHat cases, Example 1_case_mu1h=[-0_9]__mu2h=[1_1]}
        \end{subfigure}%

	  \begin{subfigure}[b]{0.45\textwidth}
                \centering
               \includegraphics[width=\textwidth]{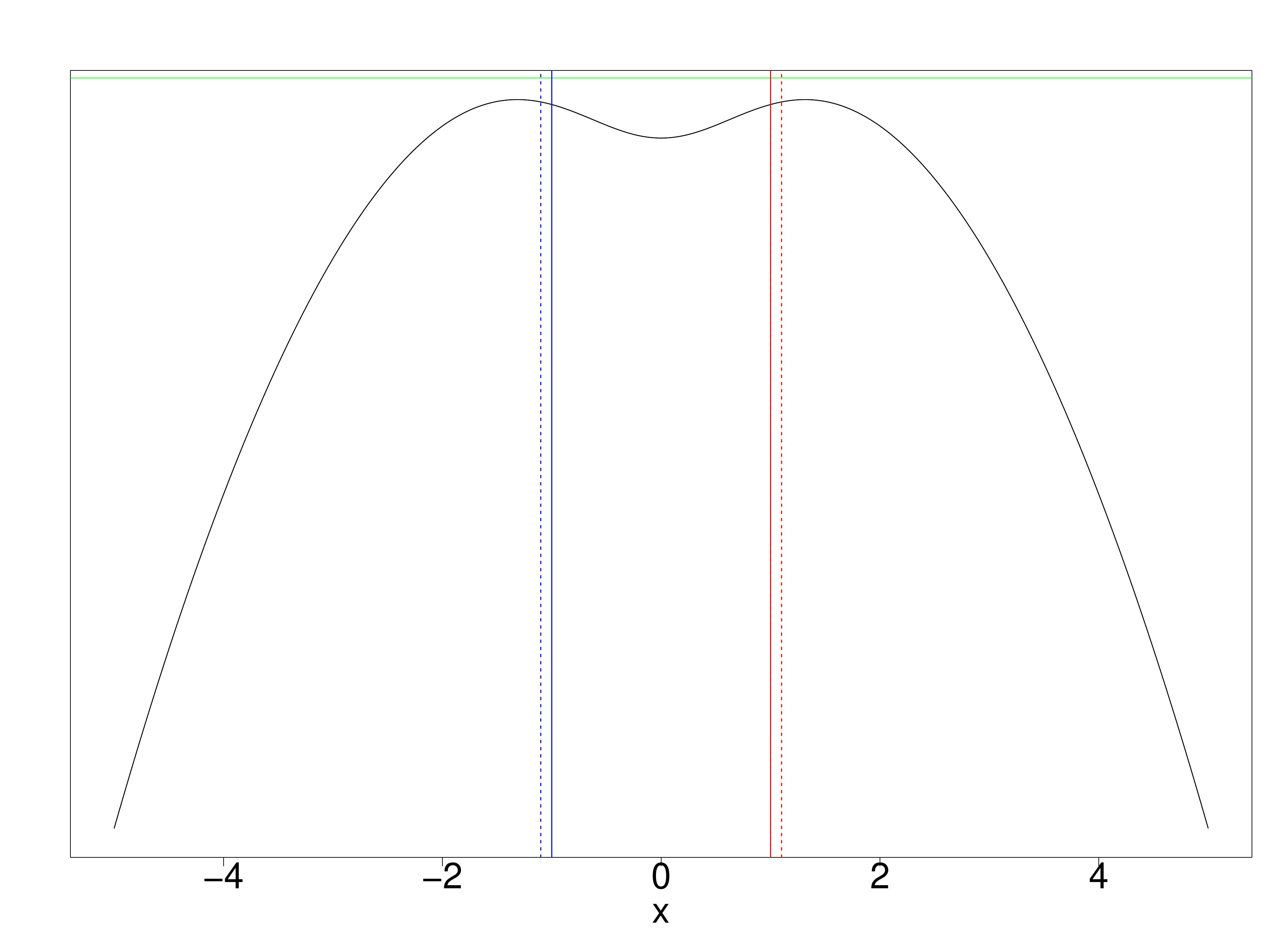}
                \caption{$\boldsymbol{\hat{\theta}} = (\hat{\mu}_1=-1.1,\hat{\mu}_2=1.1)$; \\
$\hat{\mu}_1$, $\hat{\mu}_2$ are wider, $|\hat{\mu}_j| = |{\mu}_j| + 0.1$}
                \label{fig:Qc for specific thetaHat cases, Example 1_case_mu1h=[-1_1]__mu2h=[1_1]}
        \end{subfigure}%
	  \begin{subfigure}[b]{0.45\textwidth}
                \centering
               \includegraphics[width=\textwidth]{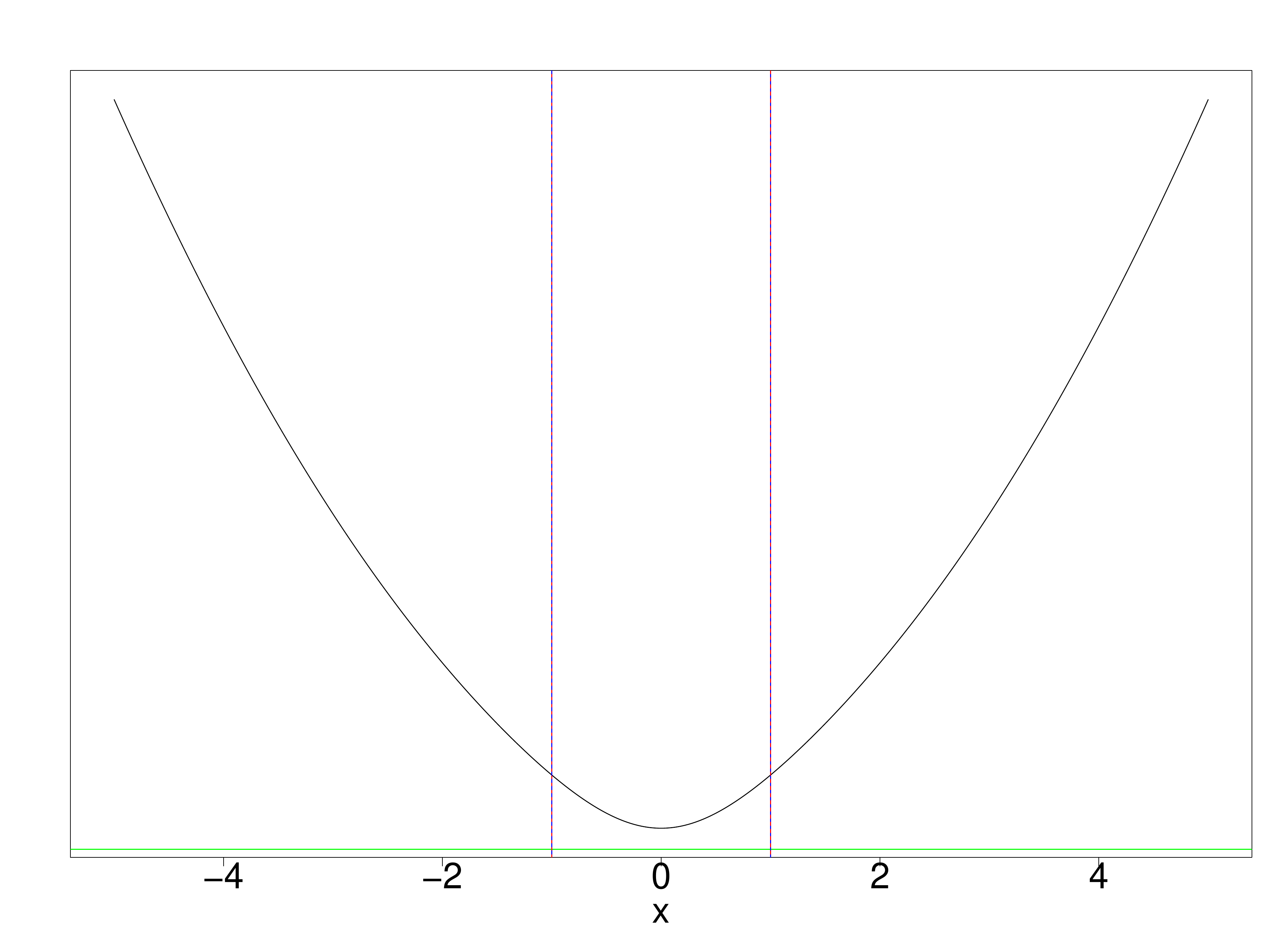}
                \caption{$\boldsymbol{\hat{\theta}} = (\hat{\mu}_1=1,\hat{\mu}_2=-1)$; \\
$\hat{\mu}_1$, $\hat{\mu}_2$ have inverse signs, $\hat{\mu}_j = -{\mu}_j$}
                \label{fig:Qc for specific thetaHat cases, Example 1_case_mu1h=[1]__mu2h=[-1]}
        \end{subfigure}%
     
        \caption{Illustration of the target $Q^c$ as a function of $x$, for specific cases of the estimated classifier parameters $\boldsymbol{\hat{\theta}} = ( \hat{\mu}_1, \hat{\mu}_2 )$.
The class mean parameters are shown in solid blue and red, with the estimated means shown in dotted blue and red. 
The green line indicates $Q^c(x) = 0$ (zero improvement); in all cases, $n_s = 18$. 
In each specific case, the optimal selection $x_*$ yields greatest correction to $\boldsymbol{\hat{\theta}}$ in terms of moving the estimated boundary $\hat{t}$ closer to the true boundary $t$.
}
	\label{fig:Qc for specific thetaHat cases, Example 1}
\end{figure}


\subsubsection{Exploration of Shannon Entropy and Random Selection}
\label{subsubsection:Exploration of Shannon Entropy and Random Selection}

The abstract example is used to compare is two selection methods, SE and RS, against optimal AL behaviour.

SE always selects $x_r$ at the estimated boundary $\hat{t}$.
RS selects uniformly from the pool, assumed to be i.i.d. in AL, hence the RS selection probability is given by the marginal density $p(x)$.
In contrast to $Q^c$ and SE, RS is a stochastic selection method, with expected selection $x_r = 0$ in this problem.
Figure \ref{fig:Se-and-RS-and-Qc for specific thetaHat cases, Example 1} illustrates $Q^c$, SE and $p(x)$ as contrasting functions of $x$, with very different maxima.

\begin{figure}
        \centering
        \begin{subfigure}[b]{0.45\textwidth}
                \centering
               \includegraphics[width=\textwidth]{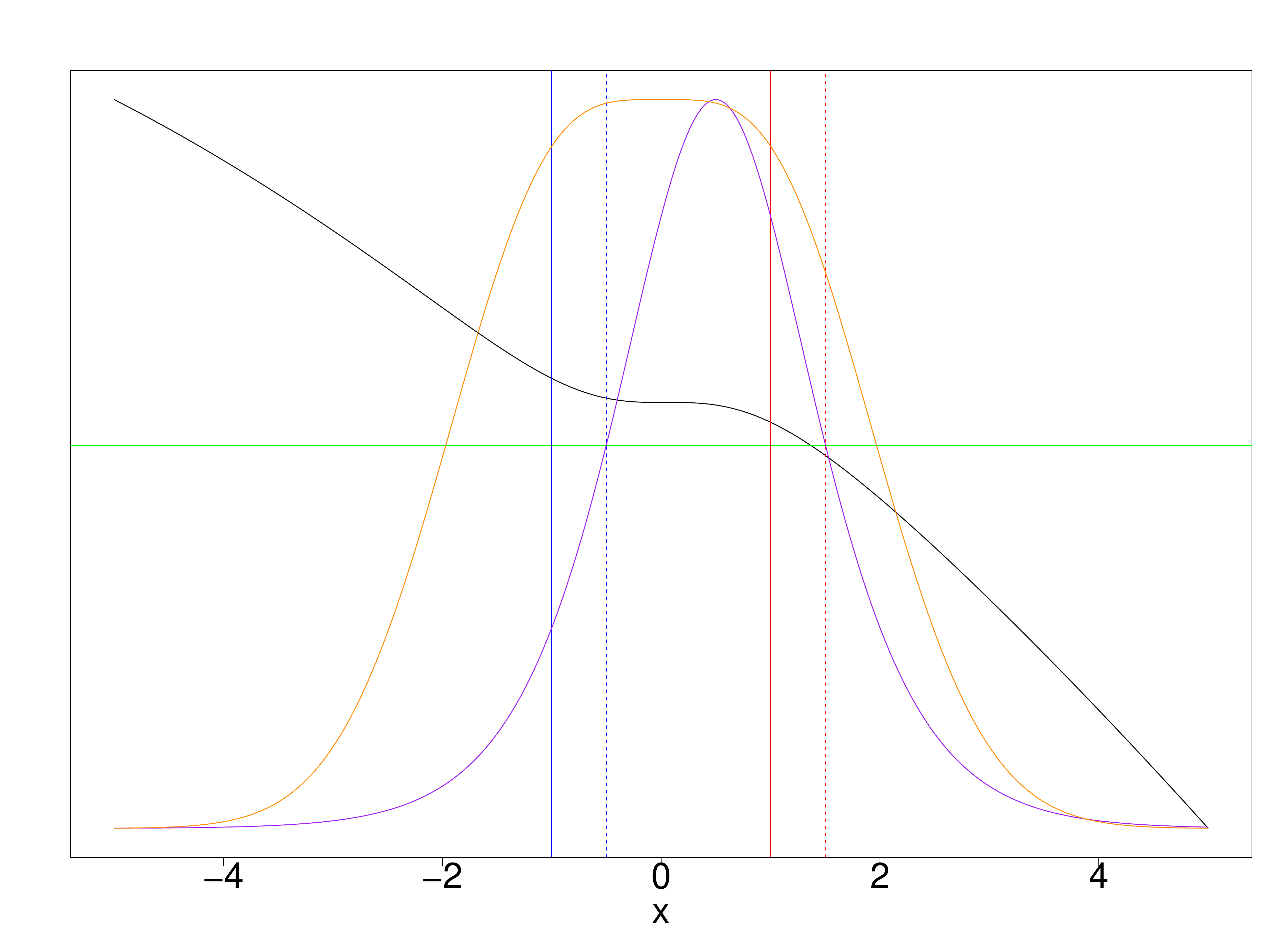}
                \caption{$\boldsymbol{\hat{\theta}} = (\hat{\mu}_1=-0.5,\hat{\mu}_2=1.5)$; \\
$\hat{\mu}_1$, $\hat{\mu}_2$ are right-shifted, $\hat{\mu}_j = {\mu}_j + 0.5$ \\}
                \label{fig:Se-and-Qc for specific thetaHat cases, Example 1_case_mu1h=[-0_5]__mu2h=[1_5]}
        \end{subfigure}
        \begin{subfigure}[b]{0.45\textwidth}
                \centering
               \includegraphics[width=\textwidth]{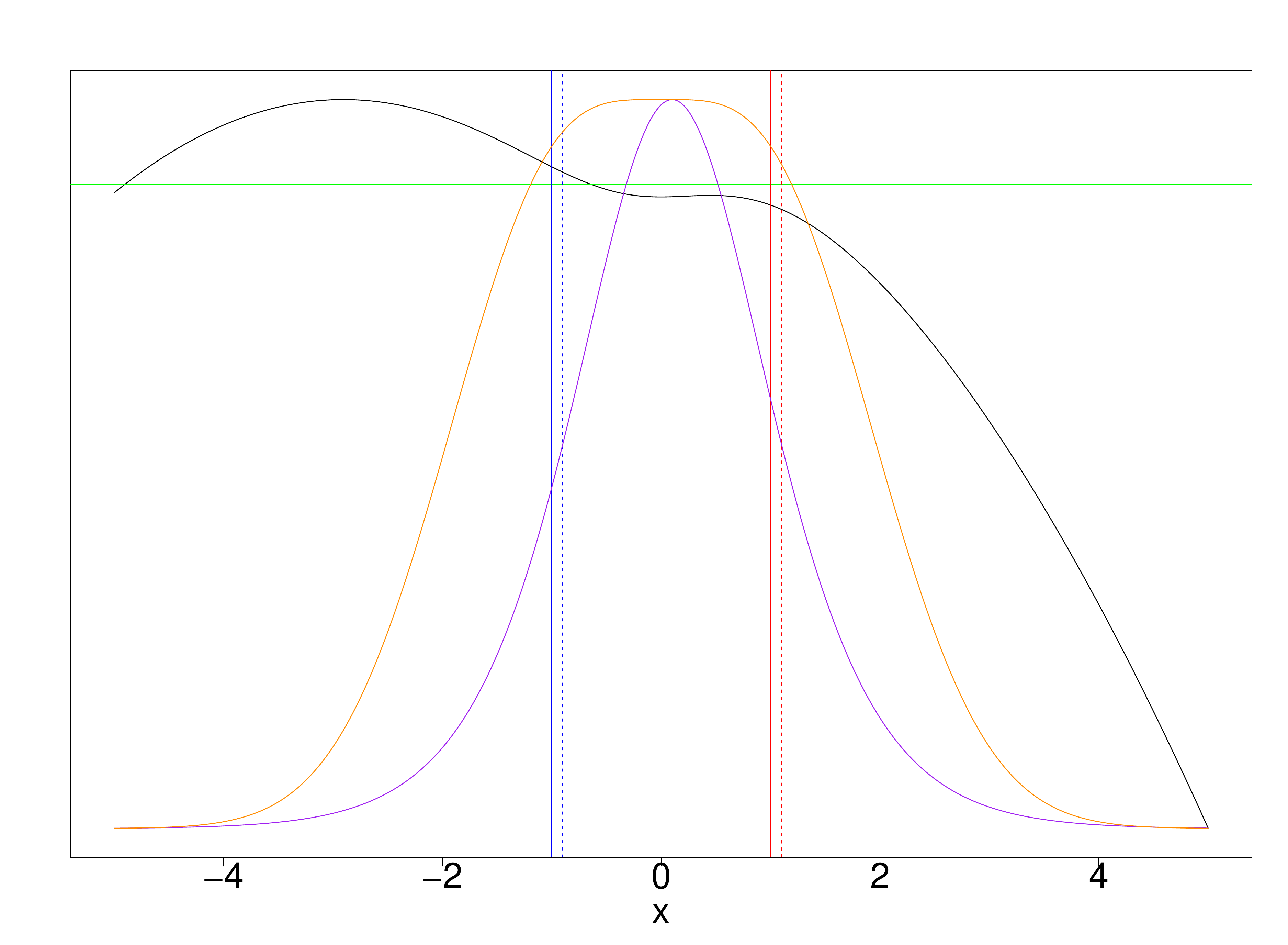}
                \caption{$\boldsymbol{\hat{\theta}} = (\hat{\mu}_1=-0.9,\hat{\mu}_2=1.1)$; \\
$\hat{\mu}_1$, $\hat{\mu}_2$ are right-shifted, $\hat{\mu}_j = {\mu}_j + 0.1$ \\}
                \label{fig:Se-and-Qc for specific thetaHat cases, Example 1_case_mu1h=[-0_9]__mu2h=[1_1]}
        \end{subfigure}%

	  \begin{subfigure}[b]{0.45\textwidth}
                \centering
               \includegraphics[width=\textwidth]{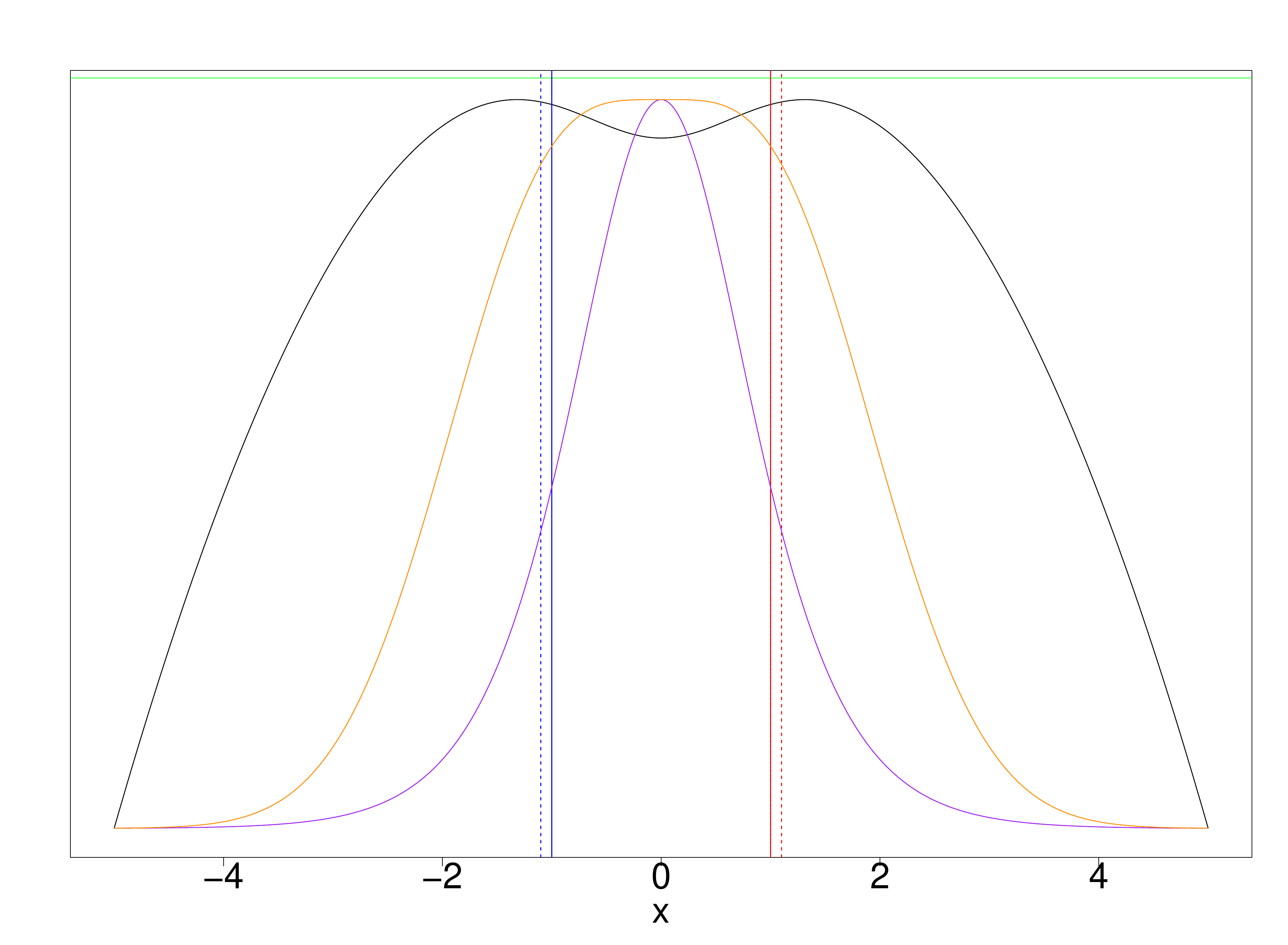}
                \caption{$\boldsymbol{\hat{\theta}} = (\hat{\mu}_1=-1.1,\hat{\mu}_2=1.1)$; \\
$\hat{\mu}_1$, $\hat{\mu}_2$ are wider, $|\hat{\mu}_j| = |{\mu}_j| + 0.1$}
                \label{fig:Se-and-Qc for specific thetaHat cases, Example 1_case_mu1h=[-1_1]__mu2h=[1_1]}
        \end{subfigure}%
	  \begin{subfigure}[b]{0.45\textwidth}
                \centering
               \includegraphics[width=\textwidth]{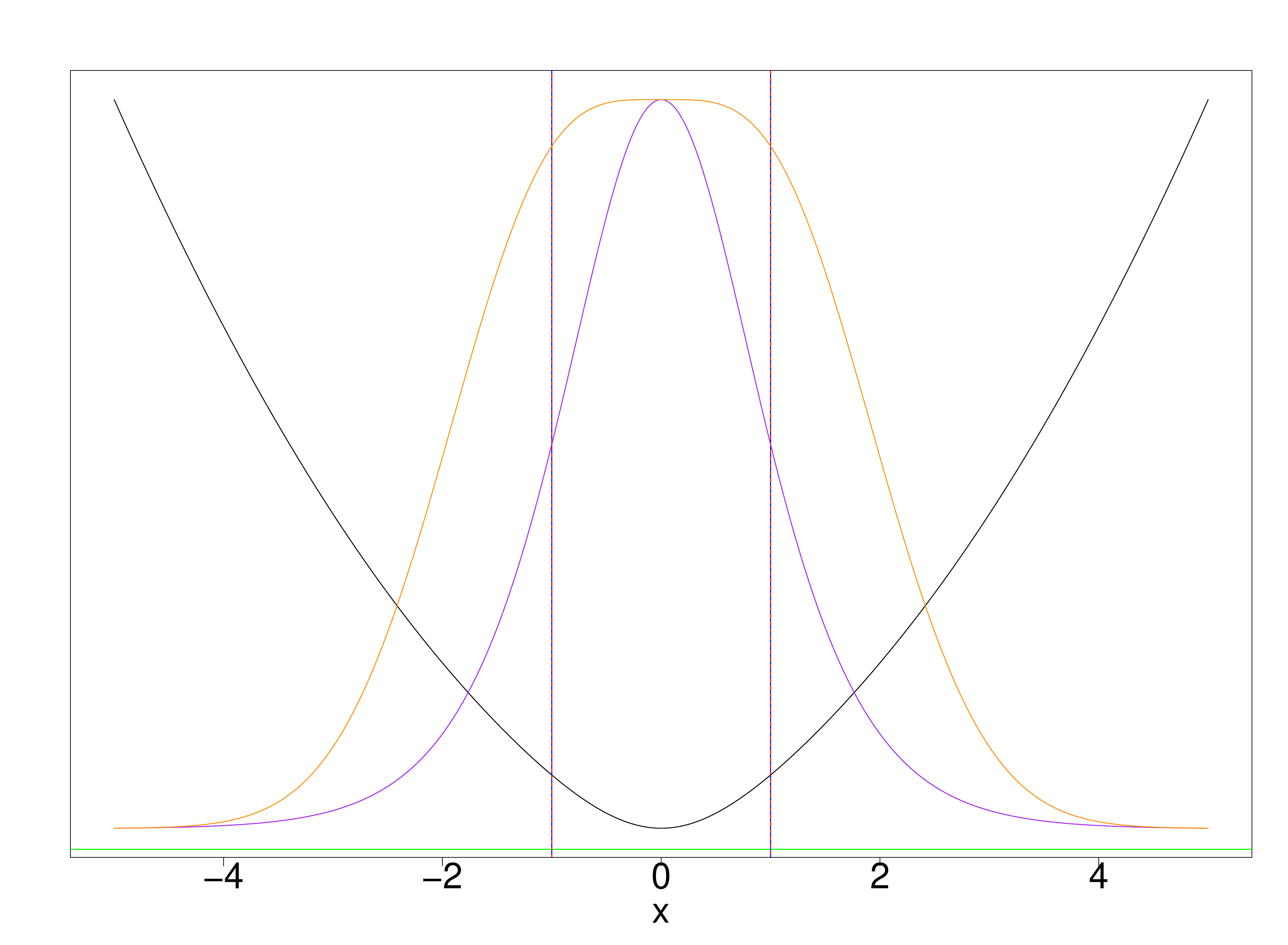}
                \caption{$\boldsymbol{\hat{\theta}} = (\hat{\mu}_1=1,\hat{\mu}_2=-1)$; \\
$\hat{\mu}_1$, $\hat{\mu}_2$ have inverse signs, $\hat{\mu}_j = -{\mu}_j$}
                \label{fig:Se-and-Qc for specific thetaHat cases, Example 1_case_mu1h=[1]__mu2h=[-1]}
        \end{subfigure}%

        \caption{Comparison of $Q^c$ against SE and RS as functions of $x$, for specific cases of the estimated classifier parameters 
$\boldsymbol{\hat{\theta}} = ( \hat{\mu}_1, \hat{\mu}_2 )$.
$Q^c$ is shown in black, SE in purple and RS in orange (for RS, the density $p(x)$ is shown).
The class mean parameters are shown in solid blue and red, with the estimated means shown in dotted blue and red. 
The green line indicates $Q^c(x) = 0$ (zero improvement); in all cases, $n_s = 18$. 
The three functions are scaled to permit this comparison.
}
	\label{fig:Se-and-RS-and-Qc for specific thetaHat cases, Example 1}
\end{figure}

$Q^c$ is asymmetric in the first two cases, and symmetric for the final two. 
By contrast, SE and RS are always symmetric (for all possible values of $\boldsymbol{\hat{\theta}}$).

In the first two cases (Figures \ref{fig:Se-and-Qc for specific thetaHat cases, Example 1_case_mu1h=[-0_5]__mu2h=[1_5]} and \ref{fig:Qc for specific thetaHat cases, Example 1_case_mu1h=[-0_9]__mu2h=[1_1]}), SE selects a central $x_r$, thereby missing the optimal selection $x_*$. 
In the second case (Figure \ref{fig:Qc for specific thetaHat cases, Example 1_case_mu1h=[-0_9]__mu2h=[1_1]}), SE selects $x_r$ with $Q^c(x_r) < 0$, failing to improve the classifier, whereas the optimal selection $x_*$ does improve the classifier since $Q^c(x_*) > 0$.
The third case is unusual, since $\hat{t} = t$ and this classifier's loss $L_e$ cannot be improved, hence $Q^c(x) < 0$ for all $x$.
In the fourth case (Figure \ref{fig:Se-and-Qc for specific thetaHat cases, Example 1_case_mu1h=[1]__mu2h=[-1]}) SE makes the worst possible choice of $x$.
In all four cases, SE never chooses the optimal point; SE may improve the classifier, but never yields the greatest improvement.
These specific cases of $\boldsymbol{\hat{\theta}}$ show that SE often makes a suboptimal choice for $x_r$, for this abstract example.

Turning to consider RS, the stochastic nature of RS suggests that the expected RS selection is the quantity of interest.
For these four cases of $\boldsymbol{\hat{\theta}}$, the expected RS selection is a suboptimal choice of $x_*$ for this problem.
It is notable that the expected RS selection is usually close to the SE selection.
The stochastic nature of RS implies that it often selects far more non-central $x$ values than SE.


\subsection{Unbiased $Q^c$ Estimation Outperforms Random Selection}
\label{subsection:Unbiased $Q^c$ Estimation Yields AL Performance Matching or Exceeding RS}

We present an argument that an unbiased estimator of $Q^c$ will always exceed RS in AL performance.
This formal approach opens the door to a new guarantee for AL, which motivates the estimation framework that MRI provides.
This guarantee is not tautological since RS generally improves the classifier, making RS a reasonable benchmark to outperform.
By contrast, heuristic AL methods such as SE lack any estimation target, making arguments of this kind difficult to construct.
This argument also motivates the algorithm bootstrapMRI (see Section \ref{subsection:Algorithm PartitionEQ}).


The context is an AL scenario with a specific classification problem, classifier and loss function.
We examine the selection of a single example from a pool $X_P$ consisting of just two examples $X_P = \{ x_1, x_2 \}$.

From Equation 2, the target function $Q^c$ depends on both the labelled data $D_S$ and the population distribution $(X,Y)$.
This dependency on both data and population is somewhat unusual for an estimation target, but other statistical targets share this property, for example classifier loss.
Here the labelled dataset $D_S$ is considered a random variable, hence the values of $Q^c$ over the pool are also random.
Consider a hypothetical $\hat{Q^c}$ estimator, unbiased in this sense: $ (\forall x_i \in \mathbb{R}) E [\hat{Q^c}(x_i, \theta, D_S)] = [Q^c(x_i, \theta, D_S)]$.

For a single example $x_i$, the true and estimated values of $Q^c$ are denoted by $Q_i = Q^c(x_i, \theta, D_S)$ and $\hat{Q}_i = \hat{Q^c}(x_i, \theta, D_S)$ respectively.
Since the estimator is unbiased, the relationship between these quantities can be conceptualised as $\hat{Q}_i = Q_i + M_i$, where $M_i$ is defined as a noise term with zero mean and variance $\sigma^2$, with $E_{D_S} [M_i] = 0$.
We assume that $M_i \condindep Q_i$, and make the moderate assumption that $M_i \sim N(0, \sigma^2)$. 

The difference between the true $Q^c$ values is defined as $R = Q_1 - Q_2$.
We begin by addressing the case where $(R > 0)$ i.e. $(Q_1 > Q_2)$.
The probability that the optimal example is chosen, denoted $\lambda$, will illustrate the estimator's behaviour under different noise variances, $\sigma^2$.

We now quantify the selection probability $\lambda$ explicitly in terms of estimator variance.
This selection probability $\lambda$ is given by 
\begin{align*}
\lambda &= p(\hat{Q_1} > \hat{Q_2}) \\ 
&= p(Q_1 + M_1 > Q_2 + M_2) \\
&= p(Q_1 - Q_2 > M_2 - M_1) \\
&= p(M_2 - M_1 < Q_1 - Q_2),
\end{align*}
which can be rewritten as $\lambda = p(N < \Delta)$ where $N= M_2 - M_1$ is defined as a mean zero RV, and $\Delta = Q_1 - Q_2$ is strictly positive (since $R > 0$).
$N$ is Gaussian, since $M_1$ and $M_2$ are both Gaussian.
This variable $\Delta$ provides a ranking signal for example selection: its sign shows that $x_1$ is a better choice than $x_2$, and its magnitude shows how much better.

Further defining $\alpha = p(N < 0)$ and $\beta = p(0 \leq N < \Delta)$ and combining with $p(N < \Delta) = p(N < 0) + p(N \leq 0 < \Delta)$ gives $\lambda = \alpha + \beta$.
Here $\alpha \condindep \Delta$ whereas $\beta \notcondindep \Delta$, showing that $\alpha$ is a pure noise term devoid of any $Q^c$ ranking information, while $\beta$ contains ranking information by its dependency on $\Delta$.

We now establish that $\alpha = \frac{1}{2}$ by examining the special case of $\hat{Q^c}$ estimator variance tending towards infinity.
This value of $\alpha = \frac{1}{2}$ proves important in relating the selection behaviour of the infinite-variance estimator to random selection.

As $\sigma^2 \uparrow \infty$, $\beta \downarrow 0$, this result being shown in Appendix A.
Hence as $\sigma^2 \uparrow \infty$, $\lambda \downarrow \alpha$.
Thus as $\sigma^2 \uparrow \infty$, $\lambda \condindep (Q_1, Q_2)$ since $\alpha \condindep (Q_1, Q_2)$, hence $\lambda$ becomes independent of true $Q^c$ values, depending only on noise. 
Hence the limiting case, as the estimator variance approaches infinity, corresponds to uniform selection over the pool.

A closely related argument for $\alpha = \frac{1}{2}$ is the impossibility of selection by signal-free noise $\alpha$ outperforming RS.
Again considering $\sigma^2 \uparrow \infty$, if $\alpha > \frac{1}{2}$ then $\lambda > \frac{1}{2}$, which will consistently prefer the better example $x_1$, and therefore consistently outperform RS.
Whereas $\alpha < \frac{1}{2}$ gives $\lambda < \frac{1}{2}$, which will consistently prefer the worse example $x_2$, and therefore consistently underperform RS.
However, outperforming RS when selecting examples by noise alone is impossible, which implies $\alpha = \frac{1}{2}$.
Further, $N$ is Gaussian with mean-zero which directly gives $\alpha = \frac{1}{2}$.

From $\alpha = \frac{1}{2}$, $\lambda$ can be expressed purely in terms of $\beta$ as
\begin{equation} \label{eq:lambda}
\lambda = \frac{1}{2} + \beta.
\end{equation}
As $\sigma^2 \uparrow \infty$, $\beta \downarrow 0$ hence $\lambda \downarrow \frac{1}{2}$.
When $\sigma^2 \downarrow 0$, $\beta \uparrow \frac{1}{2}$, as shown in Appendix B.
Hence as $\sigma^2 \downarrow 0$, $\lambda \uparrow 1$.
Since N is Gaussian, $\beta \in (0, \frac{1}{2}]$, hence $\lambda \in (\frac{1}{2}, 1]$.

Having examined the case where $(R>0)$, we now consider all of the possibilities for $R$.
The zero probability case ($R = 0$) is discarded, leaving only the second case defined by $(R < 0)$.

In this second case $(R < 0)$ i.e. $(Q_1 < Q_2)$, the optimal selection is $x_2$, with 
\begin{align*}
\lambda &= p(\hat{Q}_2 > \hat{Q}_1) \\ 
&= p(Q_2 + M_2 > Q_1 + M_1) \\
&= p(M_2 - M_1 > Q_1 - Q_2),
\end{align*}
rewritten as $\lambda = p(N > -\Delta_2)$ where $\Delta_2 = Q_2 - Q_1$ is strictly positive (since $R < 0)$.
Hence 
\begin{align*}
\lambda &= p(N > -\Delta_2) \\
&= p(N > 0) + p(-\Delta_2 < N \leq 0)  \\
&= (1 - \alpha) + \beta_2  \\
&= \frac{1}{2} + \beta_2,
\end{align*}
where $\beta_2 = p(-\Delta_2 < N \leq 0)$.
Since $N$ is Gaussian, it is symmetric, giving $\beta_2 = p(-\Delta_2 < N \leq 0) = p(0 \leq N < \Delta_2)$.

Here $\Delta$ and $\Delta_2$ differ only in magnitude, and their magnitudes do not feature in the proofs in Appendices F and G.
As a result, $\beta_2$ takes the very same values as $\beta$ when $\sigma \downarrow 0$ or $\sigma \uparrow \infty$, namely $\{ \frac{1}{2}, 0 \}$ (see the proofs in Appendices F and G).
Thus the selection behaviour of the unbiased estimator is the same for both cases of $(R>0)$ and $(R<0)$, both cases being described by Equation \ref{eq:lambda}.

The RHS of Equation \ref{eq:lambda} quantifies the combination of signal and noise in AL selection, with the estimator variance
$\sigma^2$ determining $\beta$ and $\lambda$.
Now the AL performance under the $\hat{Q^c}$ estimator can be elucidated in terms of the estimator variance.

The extreme case of infinite variance where $\lambda = \frac{1}{2}$ implies that the selection of examples is entirely random, and here the estimator's behaviour is identical to random selection (RS), an established AL benchmark.
By contrast, if $\lambda$ exceeds $\frac{1}{2}$, examples with better $Q^c$ values are more likely to be selected, leading to better expected AL performance than RS.

This argument applies directly to a pool of two elements.
The ranking of a larger pool can be decomposed into pairwise comparisons, which may extend this argument to any pool.
This argument serves to illustrate that an unbiased $\hat{Q^c}$ estimator outperforms RS, which is a new guarantee for AL.
This argument receives experimental support from the results described in Section \ref{subsection:Primary Results}.

We make no attempt to prove the existence of such an unbiased $\hat{Q^c}$ estimator.
The bootstrapMRI algorithm given in Section \ref{subsection:Algorithm PartitionEQ} is constructed, as far as is practical, to capture the key characteristics of an ideal unbiased $\hat{Q^c}$ estimator.

\section{Algorithms to Estimate Model Retraining Improvement}
\label{section:Algorithms to Estimate Example Quality}

%

For practical estimation of $\hat{Q^c}$, Term $T_c$ in Equation \ref{eq:eqc} can be ignored since it is independent of ${\bf x}$.
Thus the central task of practical $\hat{Q^c}$ estimation is the calculation of Term $T_e$ in Equation \ref{eq:eqc}, this Term $T_e$ being the expected classifier loss after retraining on the new example ${\bf x}$ with its unknown label $Y|{\bf x}$.
The definition of Term $T_e$ in Equation \ref{eq:eqc} includes two components: ${\bf p}$ and $\boldsymbol{L '}$.
Consequently, $Q^c$ estimation requires estimating these two components from one labelled dataset $D_S$.


Estimating multiple quantities from a single dataset raises interesting statistical choices.
One major choice must be made between using the same data to estimate both components (termed \emph{na\"ive reuse}), or to use bootstrapping to generate independent resampled datasets, producing independent component estimates.
This choice between na\"ive reuse and boostrapping has implications for the bias of $\hat{Q}^c$ estimates, discussed below.

Here we assume that loss estimation itself requires two datasets, for training and testing, denoted $D_T$ and $D_E$ respectively.
Since ${\bf p}$ estimation requires one dataset, then three datasets are needed in total, denoted $D_P$, $D_T$ and $D_E$, to estimate the two components ${\bf p}$ and $\mathbf{L '}$: 
\begin{itemize}
\item The class probability vector, ${\bf p} = p(Y|{\bf x})$, estimated by $\hat{\bf p}$ using dataset $D_P$,
\item The future loss vector, $\mathbf{L '}$, estimated by $\mathbf{\hat{L '}}$ using datasets $D_T$ and $D_E$.
\end{itemize}
Each of these three datasets ($D_P$, $D_T$ and $D_E$) must be derived from $D_S$.

In the case of na\"ive reuse, all three datasets equal $D_S$, yielding the algorithm simpleMRI described below.
For bootstrapping, the three datasets are all resampled from $D_S$ with replacement, giving the algorithm bootstrapMRI described below.
These two algorithms are extreme cases, chosen for clarity and performance; numerous variations are possible here.

A statistical estimate is considered \emph{precise} when it has low estimation error. 
Literature on empirical learning curves suggests that classifier loss $L$ is larger for smaller training data samples \citep{Provost2003,Gu2001,Kadie1995}.
This implies that ${\bf p}$ is difficult to estimate precisely, since precise estimates of ${\bf p}$ would directly produce a near-optimal classifier (one close to the optimum Bayes classifier, in terms of loss).
The increased loss for smaller samples further implies that loss $L$ itself is hard to estimate precisely for a small training dataset; for if loss could be precisely estimated, a near-optimal classifier could be found by direct optimisation.

This line of reasoning suggests that the two main components of $Q^c$, ${\bf p}$ and $\boldsymbol{L '}$, are both very difficult to estimate precisely from small data samples.
In practical applications where all quantities must be estimated from data, the estimates will inevitably suffer from imprecision.


\subsection{Algorithm SimpleMRI}
\label{subsection:Algorithm SimpleEQ}

We present the simpleMRI algorithm to estimate $Q^c$, to illustrate the statistical framework.
The pseudocode for simpleMRI is provided in Algorithm \ref{algorithm:SEQ}.
This first algorithm takes a simple approach where all of $D_S$ is used to estimate all three components.
The algorithm uses the maximum amount of data for each component estimate, broadly intending to reduce the variance of these component estimates.



The class probability vector $\mathbf{\hat{p}}$ is estimated by training a second classifier $\theta_2$ on $D_P$, then using its predicted probability vector $\mathbf{\hat{p}}$ for the example $\mathbf{x}$.
This second classifier is $5$-nn, or random forest when the base classifier is $k$-nn \citep{Breiman2001}.
For the future loss vector $\mathbf{\hat{L '}}$, each element ${L '}_j$ is estimated by training the base classifier $\theta()$ on $D_T \cup (\mathbf{x}, c_j)$, then computing a loss estimate using $D_E$.


The simpleMRI algorithm immediately encounters a problem in estimating Term $T_e$: the same data $D_S$ is used both to train the classifier and also to estimate the loss.
This in-sample loss estimation is known to produce optimistic, biased estimates of the loss \citep[Chapter~7]{Tibshirani2009}.
The simpleMRI algorithm suffers another potential problem with bias: the same data $D_S$ is used to estimate the class probability and estimate the loss, leading to dependence between the estimates of $\hat{\bf p}$ and $\mathbf{L '}$.
This dependence of component estimates may produce bias in the estimate $\hat{Q^c}$ from simpleMRI, since the argument of Equation \ref{eq:Tehat-unbiased} for unbiased $Q^c$ estimation requires independent component estimates.

These two problems of biased and dependent component estimates under na\"ive reuse motivates the development of a second algorithm, termed bootstrapMRI, described below. 

For computational efficiency, $\hat{Q^c}$ values are only evaluated on a randomly (uniformly) chosen subset of the pool.
This popular AL optimisation is commonly termed random sub-sampling.



\begin{algorithm}
\caption{SimpleMRI}\label{algorithm:SEQ}
\begin{algorithmic}[1]
\Procedure{SimpleMRI}{$\mathbf{x}, \theta, D_S, \theta_2$}
\State $D_P \gets D_S$ 
\State $D_T \gets D_S$
\State $D_E \gets D_S$
\BState \emph{estimate class probability vector $\mathbf{\hat{p}}$ at $\mathbf{x}$}
\State $\boldsymbol{\hat{\theta}}_2 \gets \theta_2(D_P)$
\State $\mathbf{\hat{p}} \gets \phi_2(\boldsymbol{\hat{\theta}}_2, \mathbf{x})$
\BState \emph{estimate future loss vector $\mathbf{\hat{L '}}$}
\For{$j \in [1:k]$} 
\State $\boldsymbol{\hat{\theta}}_j \gets \theta(D_T \cup (\mathbf{x}, c_j))$
\State $\hat{L '}_j \gets \frac{1}{|D_E|} \sum_{(\mathbf{x}_e, y_e) \in D_E} M_e(\boldsymbol{\hat{\theta}}_j, \mathbf{x_e}, y_e)$
\EndFor
\State $\hat{T_e} \gets \mathbf{\hat{p}} \cdot \mathbf{\hat{L '}}$
\State $\hat{Q^c} \gets \hat{T_e}$
\EndProcedure
\end{algorithmic}
\end{algorithm}

\subsection{Algorithm BootstrapMRI}
\label{subsection:Algorithm PartitionEQ}

BootstrapMRI seeks to minimise $Q^c$ estimator bias in two ways: by generating independent component estimates, and by providing component estimators of reasonably low bias.
If the two component estimators $\mathbf{\hat{p}}$ and $\mathbf{\hat{L '}}$ are independent, and both unbiased, then the $\hat{Q^c}$ estimator will be unbiased, as shown below in Section \ref{subsection:Algorithm PartitionEQ Provides Approximately Unbiased $Qc$ Estimation}.
The pseudocode for bootstrapMRI is provided in Algorithm \ref{algorithm:BEQ}.

The labelled dataset $D_S$ is resampled by bootstrapping to form three datasets $D_P, D_T$ and $D_E$.
These three datasets are independent draws from the ecdf of $D_S$, yielding independent estimates \citep[Chapter~6]{Efron1983}.

The first dataset $D_P$ provides an estimate for the class probability $\mathbf{\hat{p}}$, by classifier training on $D_P$ and class probability prediction on $\mathbf{x}$.
As before, for $\mathbf{\hat{p}}$ estimation, a second classifier $\theta_2$ is used, chosen in the very same way as simpleMRI above.
The second and third datasets $D_T$ and $D_E$ together provide an estimate of the future losses vector $\mathbf{\hat{L '}}$.
Each element ${L '}_j$ is estimated by training the base classifier $\theta()$ on $D_T \cup (\mathbf{x}, c_j)$, then computing a loss estimate using $D_E$.

In the experimental study of Section \ref{section:Experiments and Results}, the stochastic resampling is repeated, $n_b = 25$ times, and the resulting estimates are averaged.
Random sub-sampling of the pool is used for efficiency.


\begin{algorithm}
\caption{BootstrapMRI}\label{algorithm:BEQ}
\begin{algorithmic}[1]
\Procedure{BootstrapMRI}{$\mathbf{x}, \theta, D_S, n_b, \theta_2$}
\State ${\bf q} \gets \textit{zero vector of length } n_b$
\For{$b \in [1:n_b]$} 
\State $I_P \gets \textit{Sample With Replacement} (1:|D_S|)$
\State $I_T \gets \textit{Sample With Replacement} (1:|D_S|)$
\State $I_E \gets \textit{Sample With Replacement} (1:|D_S|)$
\State $D_P \gets D_S[I_P]$
\State $D_T \gets D_S[I_T]$
\State $D_E \gets D_S[I_E]$
\BState \emph{estimate class probability vector $\mathbf{\hat{p}}$ at $\mathbf{x}$}
\State $\boldsymbol{\hat{\theta}}_2 \gets \theta_2(D_P)$
\State $\mathbf{\hat{p}} \gets \phi_2(\boldsymbol{\hat{\theta}}_2, \mathbf{x})$
\BState \emph{estimate future loss vector $\mathbf{\hat{L '}}$}
\For{$j \in [1:k]$} 
\State $\boldsymbol{\hat{\theta}}_j \gets \theta(D_T \cup (\mathbf{x}, c_j))$
\State $\hat{L '}_j \gets \frac{1}{|D_E|} \sum_{(\mathbf{x}_e, y_e) \in D_E} M_e(\boldsymbol{\hat{\theta}}_j, \mathbf{x}_e, y_e)$
\EndFor
\State $\hat{T_e} \gets \mathbf{\hat{p}} \cdot \mathbf{\hat{L '}}$
\State $q[b] \gets \hat{T_e}$
\EndFor
\BState \emph{the final estimate is the average of the estimate vector $\mathbf{q}$}
\State $\hat{Q^c} \gets median(\mathbf{q})$
\EndProcedure
\end{algorithmic}
\end{algorithm}





\subsection{BootstrapMRI Algorithm Properties}
\label{subsection:Algorithm PartitionEQ Provides Approximately Unbiased $Qc$ Estimation}

BootstrapMRI seeks to minimise $\hat{Q^c}$ estimation bias by generating independent component estimates, as shown below in Equation \ref{eq:Tehat-unbiased}.
Practical $Q^c$ estimation requires calculating only Term $T_e$ in Equation 2 (Term $T_c$ can be ignored for practical estimation, since $\text{Term } T_c \condindep \mathbf{x}$).
Term $T_e$ is a product of $\mathbf{p}$ and $\mathbf{L '}$, the two components of $Q^c$ to be estimated.

The definitions of unbiased estimation are given below: 
\begin{itemize}
\item Unbiasedness for $\mathbf{\hat{p}}$ is defined as $(\forall \mathbf{x}_i \in \mathbb{R}^d) E [\mathbf{\hat{p}}(\mathbf{x}_i)] = \mathbf{p}(\mathbf{x}_i)$.
\item Unbiasedness for $\mathbf{\hat{L '}}$ is defined as $(\forall \mathbf{x}_i \in \mathbb{R}^d) E [\mathbf{\hat{L '}}(\mathbf{x}_i)] = \mathbf{L '}(\mathbf{x}_i)$.
\item Unbiasedness for $\hat{Q^c}$ is defined as $(\forall \mathbf{x}_i \in \mathbb{R}^d) E [\hat{Q^c}(\mathbf{x}_i, \theta, D_S)] = Q^c(\mathbf{x}_i, \theta, D_S)$,
\end{itemize}
where the expectations are taken over the variability of the estimators.

The independence of $\mathbf{\hat{p}}$ and $\mathbf{\hat{L '}}$ is classical statistical independence: $(\mathbf{\hat{p}} \condindep \mathbf{\hat{L '}}) 
\Leftrightarrow [ p(\mathbf{\hat{p}} = \mathbf{a}, \mathbf{\hat{L '}} = \mathbf{b}) = p(\mathbf{\hat{p}} = \mathbf{a}) \, p(\mathbf{\hat{L '}} = \mathbf{b}) ]$, for constant vectors $\mathbf{a}$ and $\mathbf{b}$.

By generating independent component estimates, bootstrapMRI provides a guarantee: that if the two component estimates $\mathbf{\hat{p}}$ and $\mathbf{\hat{L '}}$ are both unbiased, then the $\hat{Q^c}$ estimate will be unbiased.
This is shown by $E [\hat{Q^c}(x)] = Q^c(x)$, since
\begin{align} \label{eq:Tehat-unbiased}
E[\hat{T_e}] &= E[\hat{\mathbf{p}} \cdot \hat{\mathbf{L '}}] \\
&= E[\hat{\mathbf{p}}] \cdot E[\hat{\mathbf{L '}}] \nonumber \\ 
&= \mathbf{p} \cdot \mathbf{L '} \nonumber \\ 
&= T_e \nonumber.
\end{align} 

An ideal scenario would include the Bayes classifier and a large test dataset, providing the exact probabilities $\mathbf{p}$ and precise, unbiased estimates of $\mathbf{L '}$.
In that scenario, the $\hat{Q^c}$ estimate will be completely unbiased.
In the real application context, neither the Bayes classifier nor a large test dataset are available, and it is hard to estimate either component $\mathbf{\hat{p}}$ or $\mathbf{\hat{L '}}$ precisely or unbiasedly from a small data sample, these being open research problems \citep {Acharya2013,Rodriguez2013}.

Small finite samples do not permit guarantees of unbiased estimation.
In practice, the estimates of $\mathbf{\hat{p}}$ and $\mathbf{\hat{L '}}$ will suffer from both imprecision and bias.
The development of bootstrapMRI algorithm intends to approach the ideal of unbiased $\hat{Q^c}$ estimation, given the component estimators available.


For practical approximations to unbiased component estimators, we estimate $\mathbf{\hat{p}}$ and $\mathbf{\hat{L '}}$ by the $5$-nn classifier and by cross-validation respectively.
The classifier $k$-nn has well-known low asymptotic bounds on its error rate, for continuous covariates and a reasonable distance metric \citep[Chapter~6]{Ripley1996}.
These results suggest that this classifier's probability estimates should have good statistical properties, such as reasonably low bias in the finite sample case.
The estimation of $\mathbf{\hat{L '}}$ is nearly unbiased for cross-validation \citep{Efron1983}.

The class probability vector $\mathbf{p}$ is a component of $Q^c$, which raises a question for $\hat{Q^c}$ estimation, of whether $\mathbf{\hat{p}}$ estimates need to be precise for reasonable $\hat{Q^c}$ estimation.
The argument of Section \ref{subsection:Unbiased $Q^c$ Estimation Yields AL Performance Matching or Exceeding RS}, and the experimental results of bootstrapMRI in Section \ref{section:Experiments and Results}, both suggest that the $\mathbf{\hat{p}}$ estimates do not need to be very precise, but should merely have reasonably low bias.

The computational cost of EfeLc at each selection step is given by $t_a = (t_r + t_p) + (n_p \, k \, (t_r + t_l))$, where $n_p = |X_P|$ is the size of the pool, $k$ is the number of classes, $t_r$ is the cost of classifier retraining, $t_p$ is the cost of classifier prediction and $t_l$ is the cost of classifier loss estimation.
The cost for simpleMRI is the same cost as EfeLc, except that the $\hat{L}$-estimation method differs and hence $t_l$ is different.
The cost for bootstrapMRI is $n_b$ times that of simpleMRI, where $n_b$ is the number of bootstrap resamples. 

\section{Experiments and Results}
\label{section:Experiments and Results}

A large-scale experimental study explores the performance of the new $Q^c$-estimation AL methods.
The intention is to compare those methods with each other, and to standard AL methods from the literature (described in Section \ref{subsection:Literature Review}).
The focus is on the relative classifier improvements of each AL method, rather than absolute classifier performance.

The base classifier is varied, since AL performance is known to depend substantially on the classifier \citep{Guyon2011,Evans2013}.
To provide model diversity, the study uses several classifiers with different capabilities: LDA, $5$-nn, na\"ive Bayes, SVM, QDA and Logistic Regression. 
The classifiers and their implementations are described in Appendix C.

Many different classification problems are explored, including real and simulated data, described in Appendix D.
These problems are divided into three problem groups to clarify the results, see Section \ref{subsection:Assembly of Aggregate Results}.
The experimental study uses error rate for the loss function $L$ (see Section \ref{subsection:Classification}).
Further results are available for another loss function, the H measure, but are omitted for space\footnote{For these results see http://www.lewisevans.com/JMLR-Extra-Experimental-Results-Feb-2015.pdf.}; the choice of loss function does not affect the primary conclusion of Section \ref{subsection:Primary Results}.

The experimental study explores several sources of variation: the AL algorithms, the classifier $\theta$, and the classification problem $({\bf X}, Y)$.


\subsection{Active Learning Methods in the Experiment}
\label{subsection:Active Learning Methods}

The experimental study evaluates many AL methods, to compare their performance across a range of classification problems.
These methods fall into three groups: RS as the natural benchmark of AL, standard AL methods from the literature, and algorithms estimating $Q^c$.
The second group consists of four standard AL methods: SE, QbcV, QbcA, and EfeLc (all described in Section \ref{subsection:Literature Review}).
The third group contains the two $Q^c$-estimation algorithms, simpleMRI and bootstrapMRI, defined in Section \ref{section:Algorithms to Estimate Example Quality} and abbreviated as SMRI and BMRI.

For the two Qbc methods, a committee of four classifiers is chosen for model diversity: logistic regression, $5$-nn, $21$-nn, and random forest.
Random forest is a non-parametric classifier described in \citet{Breiman2001}; the other classifiers are described in Appendix C.
This committee is arbitrary, but diverse; the choices of committee size and constitution are open research problems.

Density weighting is sometimes recommended in the AL literature, see \citet{Olsson2009}.
However, the effects of density weighting are not theoretically understood. 
The experimental study also generated results from density weighting, omitted due to space\footnote{For these results see http://www.lewisevans.com/JMLR-Extra-Experimental-Results-Feb-2015.pdf.}, which left unaltered the primary conclusion that the $Q^c$-estimation algorithm bootstrapMRI is very competitive with standard methods from the literature.
The issue of density weighting is deferred to future work.

%
%


\subsection{Experimental AL Sandbox}

Iterated AL provides for the exploration of AL performance across the whole learning curve, see Section \ref{subsection:Active Learning} and \citet{Guyon2011,Evans2013}.
In this experimental study, the AL iteration continues until the entire pool has been labelled.
The pool size is chosen such that when all of the pool has been labelled, the final classifier loss is close to its asymptotic loss (that asymptotic loss being the loss from training on a much larger dataset).
The AL performance metrics described below examine the entire learning curve. 

Each single realisation of the experiment has a specific context: a classification problem, and a base classifier.
The classification data is randomly reshuffled. 
To examine variation, multiple Monte Carlo replicates are realised; ten replicates are used for each specific context.

\begin{figure}
\centering
\includegraphics[scale=0.8]{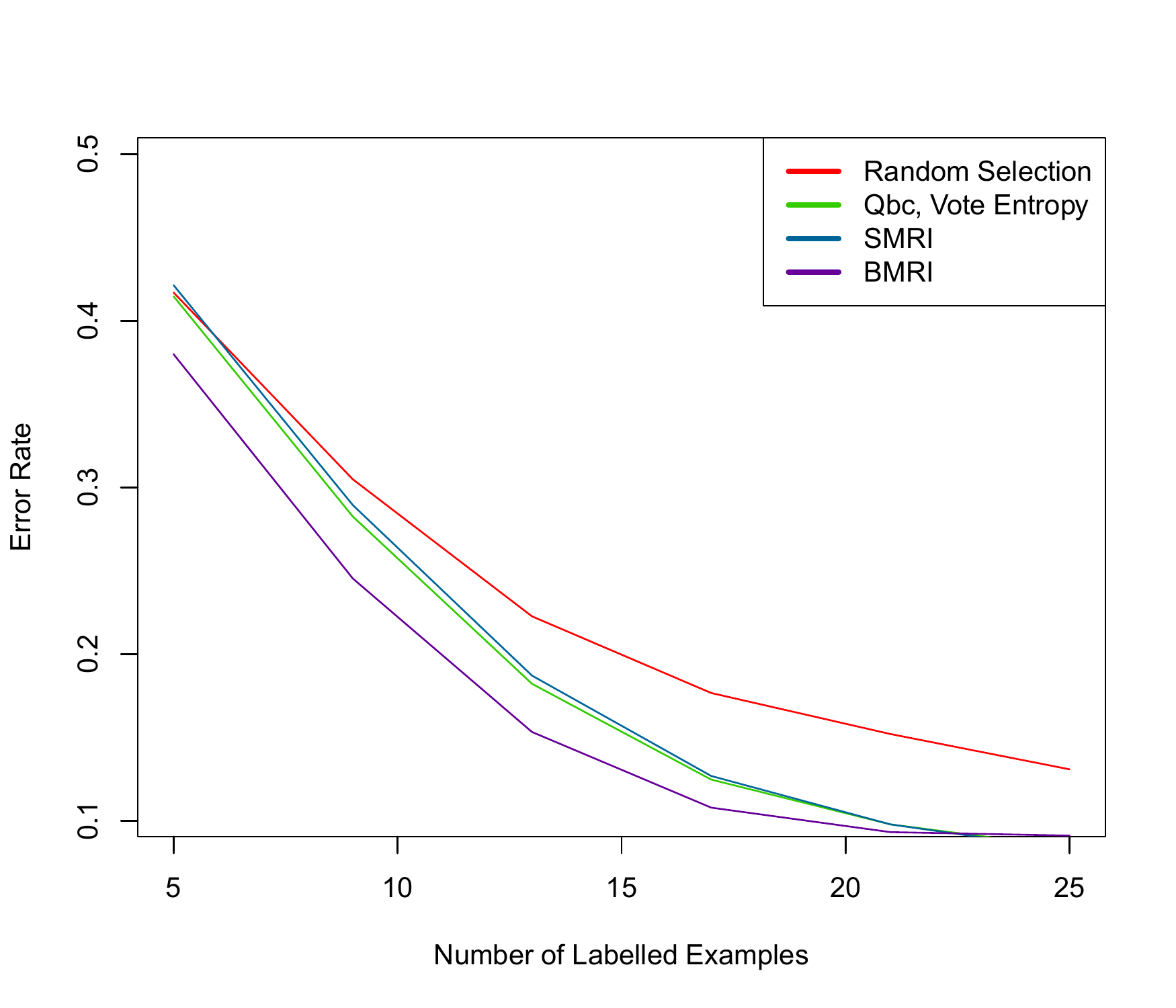}
\caption{Result for a single experiment of iterated AL. 
Each AL method performs multiple selection steps, generating a set of losses that define the learning curve. 
For clarity, a smoothed representation of the data is presented.
The early part of the learning curve is shown. 
The classification problem is the Four-Gaussian problem (see Figure \ref{fig:Ripley Four-Gaussian Problem} and Appendix D), with the base classifier being $5$-nn.
}
\label{figure:Single result for simulated data 1}
\end{figure}

Given this experimental context, the experimental AL sandbox then evaluates the performance of all AL methods over a single dataset, using iterated AL.
Each AL method produces a learning curve that shows the overall profile of loss as the number of labelled examples increases.
The amount of initial labelled data is chosen to be close to the number of classes $k$. 
To illustrate, Figure \ref{figure:Single result for simulated data 1} shows the learning curve for several AL methods, for a single realisation of the experiment.

\subsection{Assessing Performance}
\label{subsection:Assessing Performance}

As discussed in Section \ref{subsection:Active Learning}, AL performance metrics assess the relative improvements in classifier performance, when comparing one AL method against another (or when comparing AL against RS).
Thus the real quantity of interest is the ranking of the AL methods.

The AL literature provides a selection of metrics to assess AL performance, such as AUA \citep{Guyon2011}, WI \citep{Evans2013} and label complexity \citep{Dasgupta2011}.
The experimental study evaluates four metrics: AUA, WI with two weighting functions (exponential with $\alpha=0.02$, and linear), and label complexity (with 
$\epsilon=5$).
Each of these four metrics is a function of the learning curve, creating a single numeric summary from the learning curve, this curve being generated by iterated AL.

The \emph{overall rank} is also calculated as the ranking of the mean ranks, as employed, for example, by \citet{Brazdil2000}.
This yields five AL metrics in total: four primary metrics (label complexity, AUA, WI-linear, WI-exponential) and one aggregate metric (overall rank).
The overall rank avoids any arbitrary choice of one single metric.
In this experimental study, AL performance is assessed by overall rank, as used in \citet{Brazdil2000}.

For a single experiment, there is a single classification problem and base classifier.
In such an experiment, all five metrics are evaluated for every AL method, so that each metric produces its own ranking of the AL methods.
Since there are seven AL methods (see Section \ref{subsection:Active Learning Methods}), the ranks fall between one and seven, with some ties.
For brevity, the tables show the best six methods, chosen by overall rank.

The experimental results show that the AL metrics substantially agree on AL method ranking (see Tables \ref{table:Single Pairing Result-NEW 1} and \ref{table:Problem Set Result-NEW 1}).
This agreement suggests that the results are reasonably insensitive to the choice of AL metric.


\subsection{Aggregate Results}
\label{subsection:Assembly of Aggregate Results}

To address the variability of AL, multiple Monte Carlo draws are conducted for each classification problem.
Thus for each experiment, the labelled, pool and test data are drawn from the population, as different independent subsamples.
This random sampling addresses two primary sources of variation, namely the initially labelled data and the unlabelled pool.

\begin{table}[h!b!p!] 
\caption{Results for a single pair of classifier and problem, averaged over ten Monte Carlo replicates.
The base classifier is LDA. 
The classification problem is Australian Credit (see Appendix D).
These six AL methods are the best six, ordered by overall rank (calculated by numerical averages of ranks).
The $Q^c$ algorithms are shown in bold.}
\label{table:Single Pairing Result-NEW 1}
\centering
\begin{tabular}{|c|c|c|c|c|c|c|}
\hline
\multicolumn{7}{c}{Classifier LDA and Single Problem (Australian Credit)}\\
\hline
\hline
 & \textbf{BMRI} & QbcV & QbcA & EfeLc & \textbf{SMRI} & RS \\
\hline
\hline
\makecell{Overall Rank} & 1 & 2 & 3 & 4 & 5 & 6\\
\hline
\hline
\makecell{Label Complexity} & 1 & 2 & 4 & 5 & 6 & 3 \\
\hline
\makecell{AUA} & 1 & 2 & 4 & 3 & 5 & 6\\
\hline
\makecell{WI-Linear} & 2 & 1 & 3 & 5 & 4 & 6\\
\hline
\makecell{WI-Exponential} & 1 & 2 & 3 & 5 & 4 & 5\\
\hline
\end{tabular}
\end{table}
The experimental study examines many Monte Carlo draws, classification problems in groups, and classifiers.
The goal here is to determine the relative performance of the AL methods, namely to discover which methods perform better than others, on average across the whole experimental study.
To that end, the aggregate results are calculated by averaging.
First the losses are averaged, over Monte Carlo replicates.
From those losses, AL metrics are calculated, which imply overall rankings.
Finally those overall rankings are averaged, over classification problems, and then over problem groups, and finally over classifiers.
\begin{table}[h!b!p!] 
\caption{Results for a single classifier and a group of problems. 
The base classifier is QDA. 
The classification problem group is the large problem group (see Appendix D).
These six AL methods are the best six, ordered by overall rank (calculated by numerical averages of ranks).
The $Q^c$ algorithms are shown in bold.}
\label{table:Problem Set Result-NEW 1}
\centering
\begin{tabular}{|c|c|c|c|c|c|c|}
\hline
\multicolumn{7}{c}{Classifier QDA and Single Problem Group (Large Data)}\\
\hline
\hline
 & \textbf{BMRI} & \textbf{SMRI} & EfeLc & SE & RS & QbcV \\
\hline
\hline
\makecell{Overall Rank} & 1 & 2 & 3 & 4 & 5 & 6\\
\hline
\hline
\makecell{Label Complexity} & 5 & 7 & 4 & 6 & 3 & 1 \\
\hline
\makecell{AUA} & 1 & 2 & 3 & 4 & 5 & 6\\
\hline
\makecell{WI-Linear} & 2 & 1 & 3 & 4 & 5 & 6\\
\hline
\makecell{WI-Exponential} & 2 & 1 & 3 & 4 & 5 & 6\\
\hline

\end{tabular}
\end{table}

For a single pairing of classifier and problem, there are ten Monte Carlo replicates.
Consider the true distribution of AL metric scores for each AL method, where the source of the variation is the random sampling. 
The performance of each AL method is encapsulated in the score distribution, which is summarised here by the mean.
The set of mean scores implies a performance ranking of the AL methods.
These rankings are then averaged to produce a final overall ranking.
Integer rankings of the AL methods are shown for clarity.
The frequency with which each AL method outperforms random selection is also of great interest, and calculated from the group-classifier rankings.


\begin{table}[h!b!p!] 
\caption{
Results for base classifier LDA over three groups of problems.
These six AL methods are the best six, ordered by overall rank (calculated by numerical averages of ranks).
The $Q^c$ algorithms are shown in bold.}
\label{table:LDA Result-NEW Original 1}
\centering
\begin{tabular}{|c|c|c|c|c|c|c|}
\hline
\multicolumn{7}{c}{Classifier LDA}\\
\hline
\hline
\makecell{Small Problems} & QbcV & QbcA & \textbf{BMRI} & SE & \textbf{SMRI} & RS \\
\hline
\makecell{Large Problems} & SE & \textbf{BMRI} & \textbf{SMRI} & QbcA & QbcV & RS \\
\hline
\makecell{Abstract Problems} & \textbf{BMRI} & QbcV & \textbf{SMRI} & SE & RS & QbcA \\
\hline
\hline
\makecell{Average} & \textbf{BMRI} & QbcV & SE & \textbf{SMRI} & QbcA & RS\\
\hline

\end{tabular}
\end{table}

To summarise the aggregate result calculations: 
\begin{itemize}

\item[R1] For a single problem-classifier pairing, the average losses are calculated, over the ten Monte Carlo replicates.
This averaging of the losses reduces the variability in the learning curve.
From these average losses, four AL metric numbers are calculated, leading to five rankings of the AL methods, see Table \ref{table:Single Pairing Result-NEW 1}.
\item[R2] For a single group-classifier pairing, the overall rankings of all problem-classifier pairings in the group are averaged, see Table \ref{table:Problem Set Result-NEW 1}.
\item[R3] For a single classifier, the overall rankings for all three group-classifier pairings are averaged, see Table \ref{table:LDA Result-NEW Original 1} (and Tables \ref{table:KNN Result-NEW Original 1}-\ref{table:LogReg Result-NEW Original 1} in Appendix E).
\item[R4] Finally, the overall rankings for all six classifiers are averaged, see Table \ref{table:AggOverSixClassifiers Result-NEW Original 1}.
\item[R5] The frequency counts show how often each AL method outperforms RS. 
These are calculated from the group-classifier rankings (18 in total), see Table \ref{table:AggOverSixClassifiers_CompVsRs Result-NEW Original 1}.
For example, Table \ref{table:LDA Result-NEW Original 1} shows BMRI and SE outperforming RS three times out of three, whereas QbcA only outperforms RS twice.

\end{itemize}

Thus the aggregate results are calculated by averaging over successive levels, one level at a time: over Monte Carlo replicates, over problems within a group, over groups, and finally over classifiers.
This successive averaging is shown by the progression from specific realisations to the whole experiment, which starts at Figure \ref{figure:Single result for simulated data 1}, then moves through Tables \ref{table:Single Pairing Result-NEW 1} to \ref{table:AggOverSixClassifiers Result-NEW Original 1} inclusive\footnote{For further details see http://www.lewisevans.com/JMLR-Extra-Experimental-Results-Feb-2015.pdf.}.

\subsection{Results}
\label{subsection:Primary Results}

The overall performance of the AL methods is summarised by the final ranking, shown in Table \ref{table:AggOverSixClassifiers Result-NEW Original 1}, and the frequency of outperforming RS, given in Table \ref{table:AggOverSixClassifiers_CompVsRs Result-NEW Original 1}.
These two tables provide the central results for the experimental study.



\begin{table}[h!b!p!] 
\caption{
Final ranking of AL methods, over six classifiers and three groups of problems.
The $Q^c$ algorithms are shown in bold.}
\label{table:AggOverSixClassifiers Result-NEW Original 1}
\centering
\begin{tabular}{|c|c|c|c|c|c|c|c|}
\hline
\multicolumn{7}{c}{Final Ranking of AL Methods}\\
\hline
\hline
 & Rank 1 & Rank 2 & Rank 3 & Rank 4 & Rank 5 & Rank 6 & Rank 7 \\
\hline
\hline

\makecell{Average Rank} & \textbf{BMRI} & SE & QbcV & QbcA & RS & \textbf{SMRI} & EfeLc \\



\hline
\end{tabular}
\end{table}

\begin{table}[h!b!p!] 
\caption{
Frequency of outperforming RS, for six classifiers over three groups of problems.
The count shows the number of times that each AL method outperforms RS, for each group-classifier pairing (18 in total).
The count falls in the range [0,18].
The $Q^c$ algorithms are shown in bold.
}
\label{table:AggOverSixClassifiers_CompVsRs Result-NEW Original 1}
\centering
\begin{tabular}{|c|c|c|c|c|c|c|}
\hline
\multicolumn{7}{c}{Frequency of Outperforming Random Selection}\\
\hline
\hline
 & Rank 1 & Rank 2 & Rank 3 & Rank 4 & Rank 5 & Rank 6 \\
\hline
\hline
\makecell{Method} & \textbf{BMRI} & SE & QbcV & \textbf{SMRI} & QbcA & EfeLc\\
\hline
\makecell{Count better than RS} & 15 & 14 & 13 & 9 & 8 & 2\\

\hline
\end{tabular}
\end{table}

The primary conclusion is that the $Q^c$-motivated bootstrapMRI algorithm performs well in comparison to the standard AL methods from the literature.
This conclusion holds true over different classifiers and different classification problems.

Table \ref{table:AggOverSixClassifiers_CompVsRs Result-NEW Original 1} suggests that just three methods consistently outperform RS: bootstrapMRI, SE and QbcV.
BootstrapMRI outperforms RS fifteen times out of eighteen.
This provides experimental confirmation for the argument that unbiased $Q^c$ estimation algorithms consistently outperform RS, given in Section \ref{subsection:Unbiased $Q^c$ Estimation Yields AL Performance Matching or Exceeding RS}.

Comparing the $Q^c$-estimation algorithms against each other, the algorithm bootstrapMRI outperforms the algorithm simpleMRI, in all cases except two.
This suggests that minimising bias in $Q^c$ estimation may be important for AL performance.

Examining the AL methods from the literature, QBC and SE consistently perform well.
For QBC, vote entropy (QbcV) mostly outperforms average Kullback-Leibler divergence (QbcA).
EfeLc performs somewhat less well, perhaps because of the way it approximates loss using the unlabelled pool (see Section \ref{subsection:Literature Review}).
For most classifiers, RS performs poorly, with many AL methods providing a clear benefit; SVM is the exception here, where RS performs best overall.

The detailed results for each individual classifier are given in Appendix E.

Section \ref{section:Algorithms to Estimate Example Quality} describes the difficulty of estimating the $Q^c$ components, ${\bf p}$ and $\boldsymbol{L '}$, from small data samples.
For the practical algorithms, the estimates of $\boldsymbol{\hat{p}}$ and $\boldsymbol{\hat{L '}}$ will suffer from imprecision and bias.
The experimental results show that despite these estimation difficulties, strong AL performance can still be achieved.

\section{Conclusion}

Model retraining improvement is a novel statistical framework for AL, which characterises optimal behaviour via classifier loss.
This approach is both theoretical and practical, giving new insights into AL, and competitive AL algorithms for applications.

The MRI statistical estimation framework begins with the targets $Q^c$ and $B^c$.
These quantities define optimal AL behaviour for the contexts of pool-based AL: individual and batch, single-step and iterated.

Exploring the abstract definition of optimal AL behaviour generates new insights into AL.
For a particular abstract problem, the optimal selection is examined and compared to known AL methods, revealing exactly how heuristics can make suboptimal choices.
The framework is used to show that an unbiased $Q^c$ estimator will outperform random selection, bringing a new guarantee for AL.

The MRI framework motivates the construction of new algorithms to estimate $Q^c$.
A comprehensive experimental study compares the performance of $Q^c$-estimation algorithms to several standard AL methods.
The results demonstrate that bootstrapMRI is strongly competitive across a range of classifiers and problems, and is recommended for practical use.

There are many more statistical choices for $Q^c$-estimation algorithms.
These choices include various methods to estimate the class probability $\mathbf{\hat{p}}$ (e.g. via the base classifier, a different classifier, or a classifier committee); different methods to estimate the loss $\mathbf{L'}$ (e.g. in-sample, cross-validation, bootstrap, or via the unlabelled pool); and many further ways to use the data (e.g. full reuse, resampling, or partitioning).
More sophisticated estimators are the subject of future research, and hopefully MRI will motivate others to develop superior estimators.

The estimation framework enables reasoning about AL consistency behaviour and stopping rules, which are the subject of future work.
The MRI framework opens the door to potential statistical explanations of AL heuristics such as SE and Qbc, whose experimental effectiveness may otherwise remain mysterious.


\acks{
The work of Lewis P. G. Evans is supported by a doctoral training award from the EPSRC.
The authors would also like to acknowledge the valuable feedback provided by three anonymous reviewers. 
}


\appendix

\section*{Appendix A.}

This Appendix shows that given a zero-mean univariate Gaussian RV denoted $N$ with variance $\sigma$, a positive constant $\delta$, a fixed-sized interval $[0, \delta)$, and the probability $\beta = p(0 \leq N < \delta)$, then as $\sigma^2 \uparrow \infty$, $\beta \downarrow 0$.

$N$ is Gaussian with mean zero, hence it has cdf $F_N(x) = \frac{1}{2} [1 + \text{erf}(\frac{x}{\sigma \sqrt{2}})]$ where $\text{erf}(y) = \frac{1}{\sqrt{\pi}} \int^y_{-y} e^{-t^2} dt$.
By definition 
\begin{align*}
\beta &= p(0 \leq N < \delta) \\
&= F_N(\delta) - F_N(0) \\
&= F_N(\delta) - \frac{1}{2} \\
&= \frac{1}{2} [1 + \text{erf}(\frac{\delta}{\sigma \sqrt{2}})] - \frac{1}{2} \\
&= \frac{1}{2} \, \text{erf}(\frac{\delta}{\sigma \sqrt{2}}).
\end{align*}
As $\sigma^2 \uparrow \infty$, 
\begin{align*}
\text{erf}(\frac{\delta}{\sigma \sqrt{2}}) =& \frac{1}{\sqrt{\pi}} \int^{\frac{\delta}{\sigma \sqrt{2}}}_{-{\frac{\delta}{\sigma \sqrt{2}}}} e^{-t^2} dt \\
\downarrow& \frac{1}{\sqrt{\pi}} \int^0_0 e^{-t^2} dt \\
=& 0.
\end{align*}
Hence as $\sigma^2 \uparrow \infty$, $\beta \downarrow 0$.


The above argument applies to a RV $N$ and a fixed interval $[0, \delta)$, but also applies to a random interval $[0, \Delta)$ with $\Delta$ being a strictly positive RV, since the argument relies to every realisation of $\Delta$.

\section*{Appendix B.}

This Appendix shows that given a zero-mean univariate Gaussian RV denoted $N$ with variance $\sigma$, a positive constant $\delta$, a fixed-sized interval $[0, \delta)$, and the probability $\beta = p(0 \leq N < \delta)$, then as $\sigma^2 \downarrow 0$, $\beta \uparrow \frac{1}{2}$.

By definition, 
\begin{align*}
p(N \geq 0) = p(0 \leq N < \delta) + p(N \geq \delta), 
\end{align*}
i.e.
\begin{align*}
\frac{1}{2} = \beta + p(N \geq \delta),
\end{align*}
giving
\begin{align*}
\beta = \frac{1}{2} - p(N \geq \delta).
\end{align*}

By definition 
\begin{align*}
p(N \geq \delta) \leq p(|N| \geq \delta),
\end{align*} 
and Chebyshev's Inequality gives 
\begin{align*}
p(|N| \geq \delta) \leq \frac{\sigma^2}{\delta^2},
\end{align*}
hence 
\begin{align*}
p(N \geq \delta) \leq \frac{\sigma^2}{\delta^2}.
\end{align*}
As $\sigma^2 \downarrow 0$, $\frac{\sigma^2}{\delta^2} \downarrow 0$ hence $p(N \geq \delta) \downarrow 0$.
Thus as $\sigma^2 \downarrow 0$, $p(N \geq \delta) \downarrow 0$, combining with $\beta = \frac{1}{2} - p(N \geq \delta)$ yields $\beta \uparrow \frac{1}{2}$ as $\sigma^2 \downarrow 0$.


The above argument applies to a RV $N$ and a fixed interval $[0, \delta)$, but also applies to a random interval $[0, \Delta)$ with $\Delta$ being a strictly positive RV, since the argument relies to every realisation of $\Delta$.

\section*{Appendix C.}

This Appendix describes the six classifiers used in the experimental study of Section \ref{section:Experiments and Results}, and their implementation details.

Section \ref{section:Experiments and Results} describes experiments with six classifiers: linear discriminant analysis, quadratic discriminant analysis, $K$-nearest-neighbours, na\"ive Bayes, logistic regression and support vector machine.
Linear discriminant analysis (LDA) is a linear generative classifier described in \citet[Chapter~4]{Tibshirani2009}.
Quadratic discriminant analysis (QDA) is a non-linear generative classifier described in \citet[Chapter~4]{Tibshirani2009}.
$K$-Nearest-Neighbours ($K$-nn) is a well-known non-parametric classifier discussed in \citet[Chapter~4]{Duda2001}.
Na\"ive Bayes is a probabilistic classifier which assumes independence of the covariates, given the class; see \citet{Hand2001}.
Logistic Regression is a linear parametric discriminative classifier described in \citet[Chapter~4]{Tibshirani2009}.
The support vector machine (SVM) is a popular non-parametric classifier described in \citet{Vapnik1995}. 
Standard R implementations are used for these classifiers, see below. 

The classifier implementation details are now described.
For LDA, the standard R implementation is used.
For QDA, the standard R implementation is used.
For $5$-nn, the R implementation from package kknn is used.\footnote{For details see http://cran.r-project.org/web/packages/kknn/kknn.pdf.}
This implementation applies covariate scaling: each covariate is scaled to have equal standard deviation (using the same scaling for both training and testing data).
For na\"ive Bayes, the R implementation from package e1071 is used.\footnote{For details see http://cran.r-project.org/web/packages/e1071/e1071.pdf.}
For continuous predictors, a Gaussian distribution (given the target class) is assumed.
This approach is less than ideal, but tangential to the statistical estimation framework and experimental study.
For Logistic Regression, the Weka implementation from package RWeka is used.\footnote{For details see http://cran.r-project.org/web/packages/RWeka/RWeka.pdf.}
For SVM, the R implementation from package e1071 is used.
The SVM kernel used is radial basis kernel.
The probability calibration of the scores is performed for binary problems by MLE fitting of a logistic distribution to the decision values, or for multi-class problems, by computing the a-posteriori class probabilities using quadratic optimisation.


\section*{Appendix D.}

A diverse set of classification problems is chosen to explore AL performance.
The classification problems fall into two sets: real problems and abstract problems.

First the real data classification problems are shown in Tables \ref{table:Small Real Data 1} and \ref{table:Large Real Data 1}.
The real data problems are split into two groups, one for smaller problems of fewer examples, and another of larger problems.
The class prior is shown, since the experimental study uses error rate as loss.
The sources for this data include UCI \citep{Bache2013}, \citet{Guyon2011}, \citet{Anagnostopoulos2012} and \citet{Adams2010}.

The intention here is to provide a wide variety in terms of problem properties: covariate dimension $d$, number of classes $k$, the class prior $\boldsymbol{\pi}$, and the underlying distribution.
The number and variety of problems suggests that the results in Section \ref{section:Experiments and Results} have low sensitivity to the presence or absence of one or two specific problems.

\begin{table}[h!b!p!] 
\caption{Real Data Classification Problems, Smaller}
\label{table:Small Real Data 1}
\centering
\begin{tabular}{|c|c|c|c|c|}
\hline
Name & Dim. $d$ & Classes $k$ & Cases $n$ & Class Prior $\boldsymbol{\pi}$ \\
\hline
\hline
\makecell{Australian} & 14 & 2 & 690 & (0.44, 0.56)\\
\hline
\makecell{Balance} & 4 & 3 & 625 & (0.08, 0.46, 0.46)\\
\hline
\makecell{Glass} & 10 & 6 & 214 & (0.33,0.36,0.08,0.06,0.04,0.14)\\
\hline
\makecell{Heart-Statlog} & 13 & 2 & 270 & (0.65, 0.44)\\
\hline
\makecell{Monks-1} & 6 & 2 & 432 & (0.5, 0.5)\\
\hline
\makecell{Monks-2} & 6 & 2 & 432 & (0.5, 0.5)\\
\hline
\makecell{Monks-3} & 6 & 2 & 432 & (0.5, 0.5)\\
\hline
\makecell{Pima Diabetes} & 8 & 2 & 768 & (0.35, 0.65)\\
\hline
\makecell{Sonar} & 60 & 2 & 208 & (0.47, 0.53)\\
\hline
\makecell{Wine} & 13 & 3 & 178 & (0.33, 0.4, 0.27)\\
\hline
\end{tabular}
\end{table}


\begin{table}[h!b!p!] 
\caption{Real Data Classification Problems, Larger}
\label{table:Large Real Data 1}
\centering
\begin{tabular}{|c|c|c|c|c|}
\hline
Name & Dim. $d$ & Classes $k$ & Cases $n$ & Class Prior $\boldsymbol{\pi}$ \\
\hline
\hline
\makecell{Fraud} & 20 & 2 & 5999 & (0.167, 0.833)\\
\hline
\makecell{Electricity Prices} & 6 & 2 & 27552 & (0.585, 0.415)\\ 
\hline
\makecell{Colon} & 16 & 2 & 17076 & (0.406, 0.594)\\ 
\hline
\makecell{Credit 93} & 29 & 2 & 4406 & (0.007, 0.993)\\ 
\hline
\makecell{Credit 94} & 29 & 2 & 8493 & (0.091, 0.909)\\ 
\hline
\end{tabular}
\end{table}


Second the abstract classification problems are illustrated in Figure \ref{fig:Contour graphs to show the tasks}.
These abstract problems are generated by sampling from known probability distributions.
The class-conditional distributions $({\bf X} | y = c_j)_1^k$ are either Gaussians or mixtures of Gaussians. 
This set of problems presents a variety of decision boundaries to the classifier.
All have balanced uniform priors, and the Bayes Error Rates are approximately 0.1.


\def \Sim_Problem_Textwidth_Multiplier {0.19}

\begin{figure}
        \centering
        \begin{subfigure}[b]{\Sim_Problem_Textwidth_Multiplier \textwidth}
                \centering
                \includegraphics[width=\textwidth]{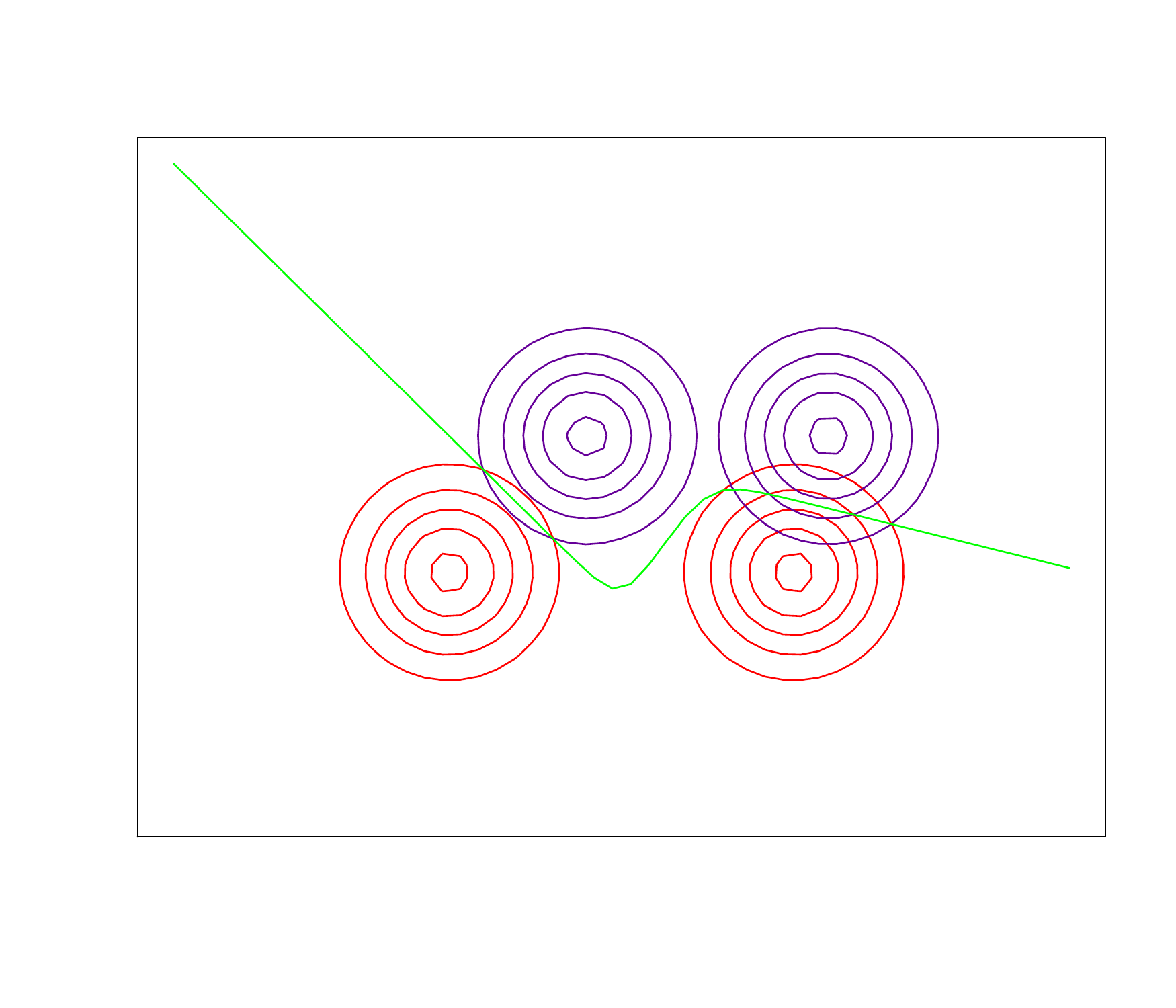} 
                \caption{Four- \\ Gaussian, \\ see \citep{Ripley1996}}
                \label{fig:Ripley Four-Gaussian Problem}
        \end{subfigure}
        \begin{subfigure}[b]{\Sim_Problem_Textwidth_Multiplier \textwidth}
                \centering
                \includegraphics[width=\textwidth]{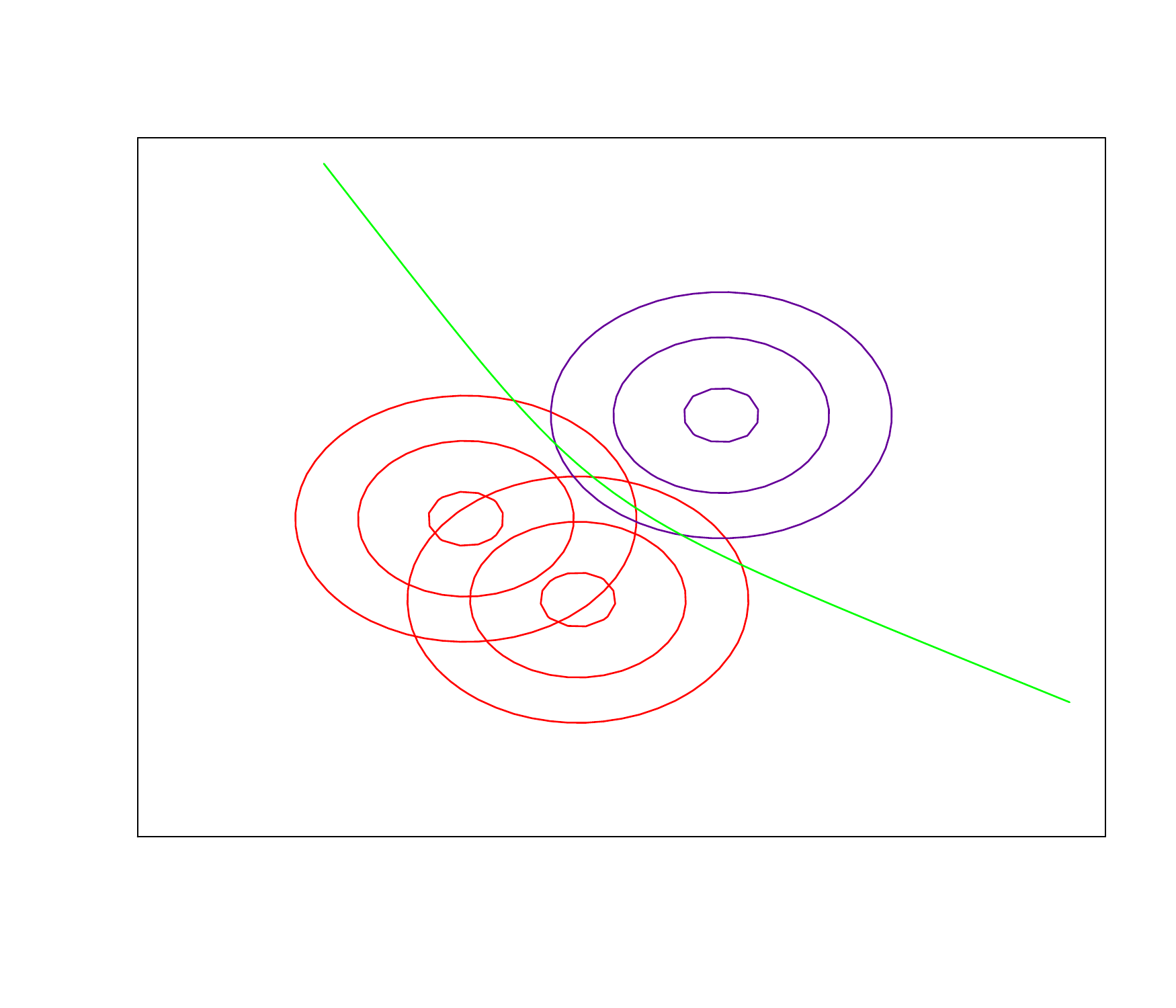} 
                \caption{Gaussian \\ Quadratic \\ boundary}
        \end{subfigure}%
        \begin{subfigure}[b]{\Sim_Problem_Textwidth_Multiplier \textwidth}
                \centering
                \includegraphics[width=\textwidth]{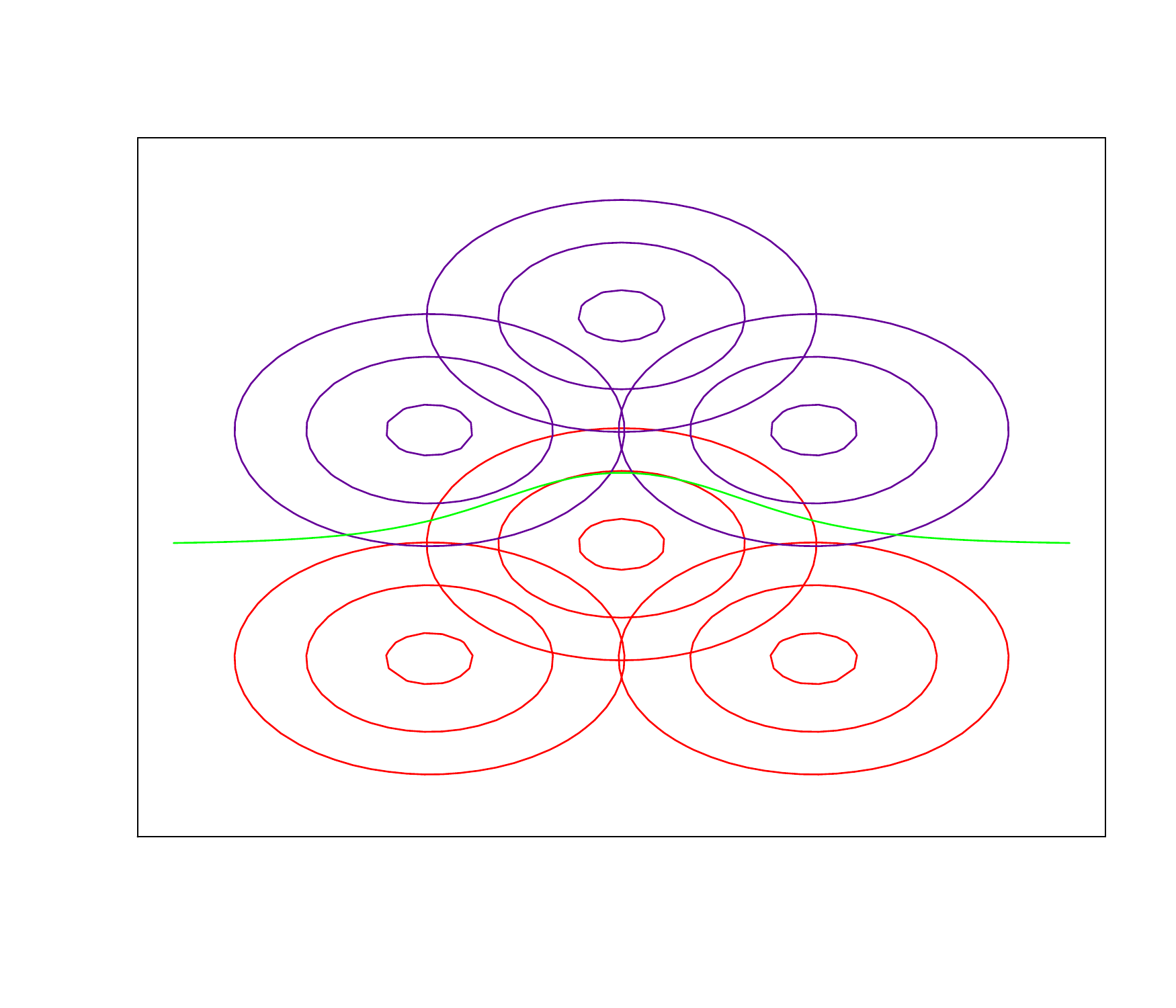} 
                \caption{Triangles of \\ three Gaussians\\}
        \end{subfigure}
        \begin{subfigure}[b]{\Sim_Problem_Textwidth_Multiplier \textwidth}
                \centering
                \includegraphics[width=\textwidth]{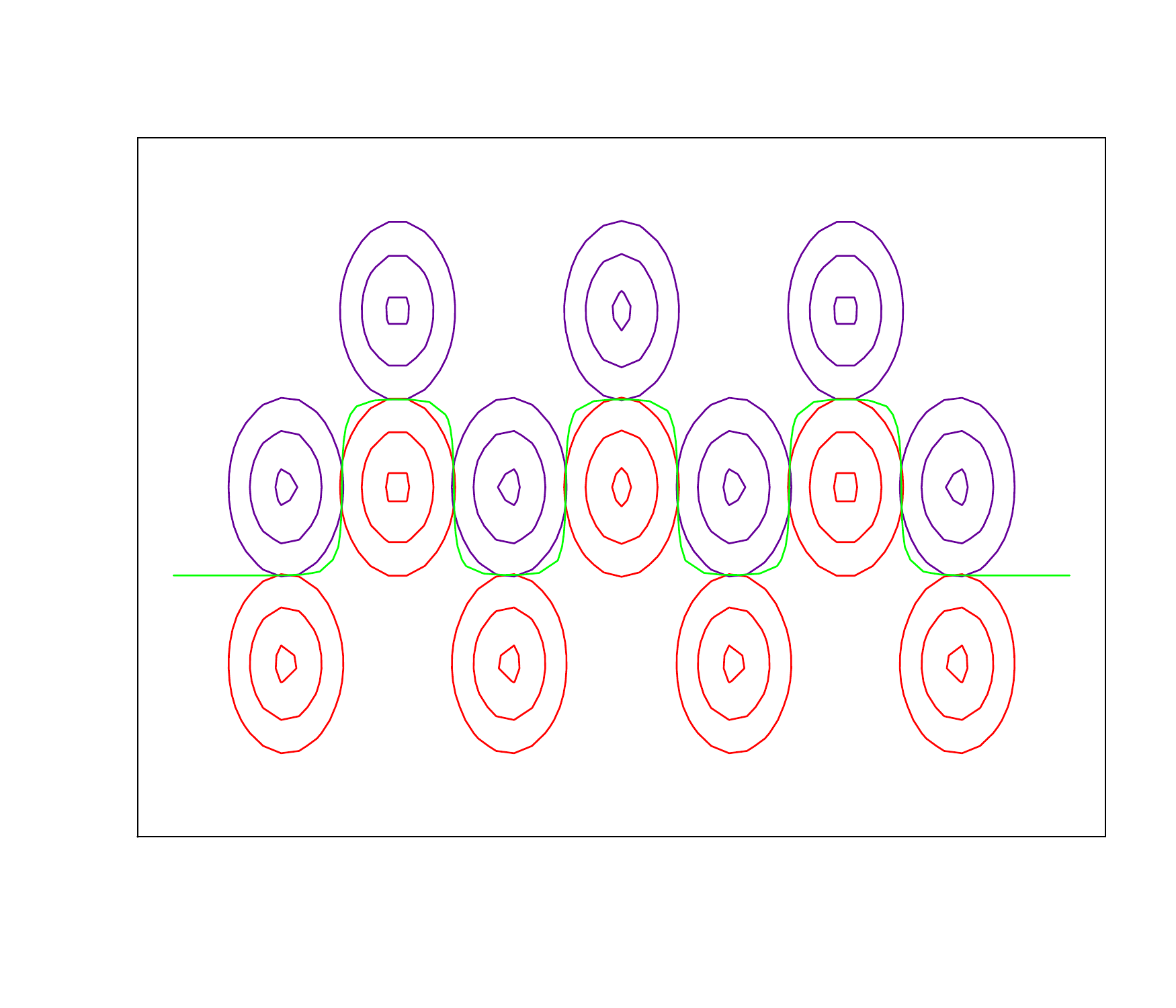} 
                \caption{Gaussian sets\\ oscillating-\\boundary}
        \end{subfigure}
        \centering
        \begin{subfigure}[b]{\Sim_Problem_Textwidth_Multiplier \textwidth}
                \centering
                \includegraphics[width=\textwidth]{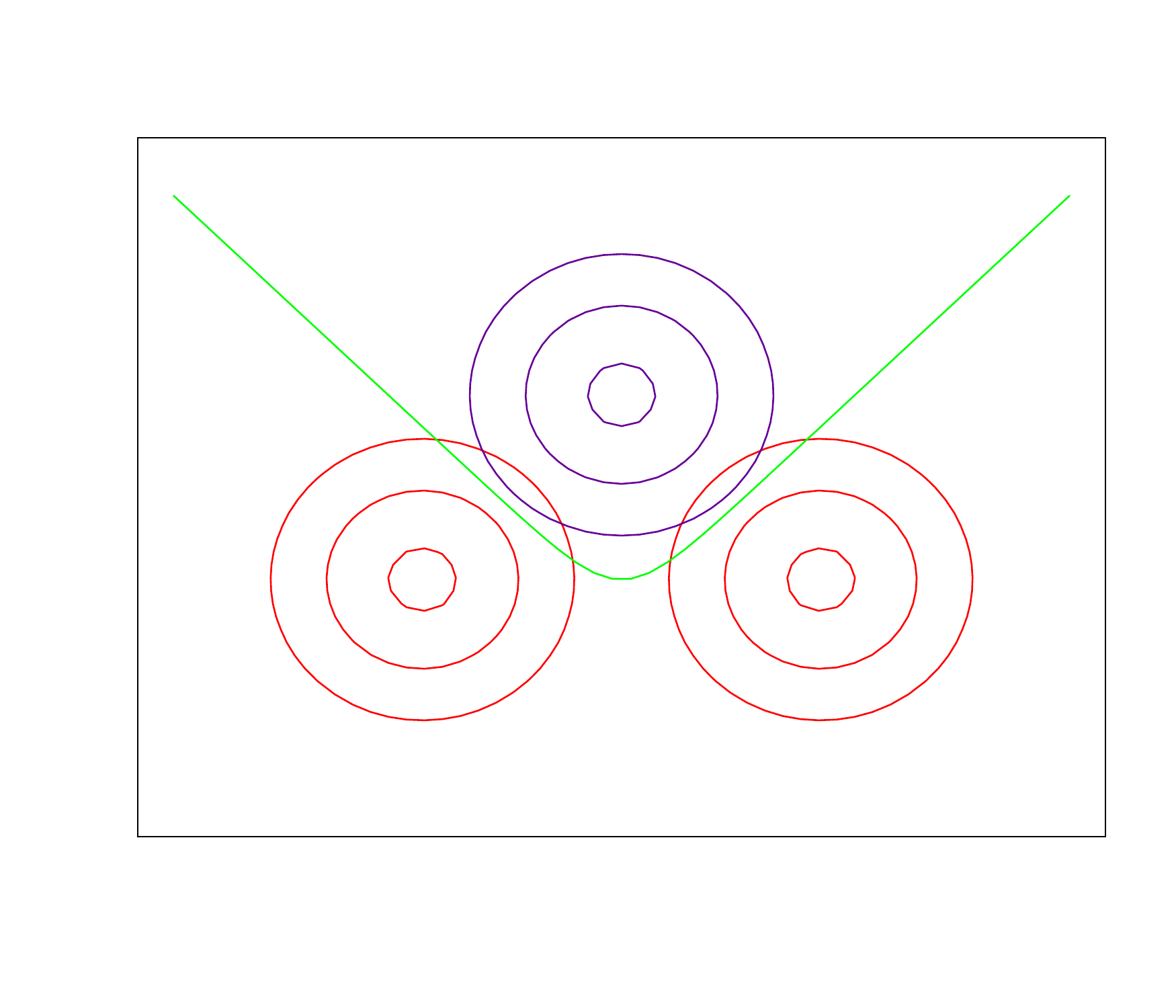} 
                \caption{Gaussian \\ sharply non-linear\\boundary} 
        \end{subfigure}
     
        \caption{Density contour plots showing the abstract classification problems. 
The class-conditional distributions $({\bf X} | y = c_j)_1^k$ are shown in red for class 1 and blue for class 2.
These class-conditional distributions $({\bf X} | y = c_j)_1^k$ are either Gaussians or mixtures of Gaussians.
The decision boundary is shown in green.
}
	  \label{fig:Contour graphs to show the tasks}
\end{figure}

\section*{Appendix E.}

This Appendix shows the results for each individual classifier in the experimental study described in Section \ref{section:Experiments and Results}.
The results for LDA, $K$-nn, na\"ive Bayes, SVM, QDA and Logistic Regression are shown in Tables \ref{table:LDA Result-NEW Original 1}, \ref{table:KNN Result-NEW Original 1}, \ref{table:Naive Bayes Result-NEW Original 1}, \ref{table:SVM Result-NEW Original 1}, \ref{table:QDA Result-NEW Original 1} and \ref{table:LogReg Result-NEW Original 1} respectively.
These results are the detailed results of the experimental study, covering the six classifiers, all the problems in three groups, and multiple Monte Carlo replicates.
In each table, the average rank is calculated as the numerical mean, with ties resolved by preferring lower variance of the rank vector.


\begin{table}[h!b!p!] 
\caption{
Results for base classifier $5$-nn over three groups of problems.
These six AL methods are the best six, ordered by overall rank (calculated by numerical averages of ranks).
The $Q^c$ algorithms are shown in bold.}
\label{table:KNN Result-NEW Original 1}
\centering
\begin{tabular}{|c|c|c|c|c|c|c|}
\hline
\multicolumn{7}{c}{Classifier $5$-nn}\\
\hline
\hline
\makecell{Small Problems} & SE & \textbf{BMRI} & QbcV & \textbf{SMRI} & RS & QbcA \\
\hline
\makecell{Large Problems} & QbcA & \textbf{BMRI} & SE & QbcV & RS & \textbf{SMRI} \\ 
\hline
\makecell{Abstract Problems} & \textbf{BMRI} & SE & RS & QbcV & \textbf{SMRI} & QbcA \\
\hline
\hline
\makecell{Average} & \textbf{BMRI} & SE & QbcV & RS & QbcA & \textbf{SMRI}\\
\hline
\end{tabular}
\end{table}

\begin{table}[h!b!p!] 
\caption{
Results for base classifier na\"ive Bayes over three groups of problems.
These six AL methods are the best six, ordered by overall rank (calculated by numerical averages of ranks).
The $Q^c$ algorithms are shown in bold.}
\label{table:Naive Bayes Result-NEW Original 1}
\centering
\begin{tabular}{|c|c|c|c|c|c|c|}
\hline
\multicolumn{7}{c}{Classifier na\"ive Bayes}\\
\hline
\hline
\makecell{Small Problems} & SE & \textbf{BMRI} & QbcV & QbcA & RS & \textbf{SMRI} \\
\hline
\makecell{Large Problems} & QbcV & SE & EfeLc & \textbf{BMRI} & \textbf{SMRI} & QbcA \\
\hline
\makecell{Abstract Problems} & SE & QbcV & \textbf{BMRI} & \textbf{SMRI} & RS & QbcA \\
\hline
\hline
\makecell{Average} & SE & QbcV & \textbf{BMRI} & \textbf{SMRI} & QbcA & RS\\
\hline
\end{tabular}
\end{table}

\begin{table}[h!b!p!] 
\caption{
Results for base classifier SVM over three groups of problems.
These six AL methods are the best six, ordered by overall rank (calculated by numerical averages of ranks).
The $Q^c$ algorithms are shown in bold.}
\label{table:SVM Result-NEW Original 1}
\centering
\begin{tabular}{|c|c|c|c|c|c|c|}
\hline
\multicolumn{7}{c}{Classifier SVM}\\
\hline
\hline
\makecell{Small Problems} & QbcV & RS & QbcA & SE & \textbf{BMRI} & \textbf{SMRI} \\
\hline
\makecell{Large Problems} & RS & QbcA & QbcV & EfeLc & \textbf{SMRI} & \textbf{BMRI} \\
\hline
\makecell{Abstract Problems} & QbcV & RS & QbcA & \textbf{BMRI} & \textbf{SMRI} & SE \\
\hline
\hline
\makecell{Average} & RS & QbcV & QbcA & \textbf{BMRI} & \textbf{SMRI} & SE \\
\hline
\end{tabular}
\end{table}

\begin{table}[h!b!p!] 
\caption{
Results for base classifier QDA over three groups of problems.
These six AL methods are the best six, ordered by overall rank (calculated by numerical averages of ranks).
The $Q^c$ algorithms are shown in bold.}
\label{table:QDA Result-NEW Original 1}
\centering
\begin{tabular}{|c|c|c|c|c|c|c|}
\hline
\multicolumn{7}{c}{Classifier QDA}\\
\hline
\hline
\makecell{Small Problems} & SE & \textbf{BMRI} & QbcV & QbcA & \textbf{SMRI} & RS \\
\hline
\makecell{Large Problems} & \textbf{BMRI} & \textbf{SMRI} & EfeLc & SE & RS & QbcV \\
\hline
\makecell{Abstract Problems} & SE & \textbf{BMRI} & RS & QbcV & QbcA & \textbf{SMRI} \\
\hline
\hline
\makecell{Average} & \textbf{BMRI} & SE & QbcV & \textbf{SMRI} & RS & QbcA \\
\hline
\end{tabular}
\end{table}

\begin{table}[h!b!p!] 
\caption{
Results for base classifier Logistic Regression over three groups of problems.
These six AL methods are the best six, ordered by overall rank (calculated by numerical averages of ranks).
The $Q^c$ algorithms are shown in bold.}
\label{table:LogReg Result-NEW Original 1}
\centering
\begin{tabular}{|c|c|c|c|c|c|c|}
\hline
\multicolumn{7}{c}{Classifier Logistic Regression}\\
\hline
\hline
\makecell{Small Problems} & QbcV & QbcA & \textbf{BMRI} & SE & RS & \textbf{SMRI} \\
\hline
\makecell{Large Problems} & SE & QbcV & QbcA & \textbf{SMRI} & \textbf{BMRI} & RS \\
\hline
\makecell{Abstract Problems} & \textbf{BMRI} & RS & SE & \textbf{SMRI} & QbcV & QbcA \\
\hline
\hline
\makecell{Average} & QbcV & SE & \textbf{BMRI} & QbcA & RS & \textbf{SMRI} \\
\hline
\end{tabular}
\end{table}

The results of Section \ref{subsection:Primary Results} quantify the benefit of AL over RS: the rankings of Tables \ref{table:LDA Result-NEW Original 1} and \ref{table:AggOverSixClassifiers Result-NEW Original 1} show how much AL methods outperform RS.
Another way to quantify AL benefit is provided by the regret difference between an AL method and RS.
Here AL regret is naturally defined as the loss difference, between the optimal performance given by maximising $Q^c$, and the actual performance of any given AL method.
Another aspect of AL benefit is the question of where AL outperforms RS, and this aspect is quantified by the frequency results in Table \ref{table:AggOverSixClassifiers_CompVsRs Result-NEW Original 1}.

\vskip 0.2in

\bibliography{Shared}

\end{document}